\pdfoutput=1

\documentclass[11pt]{article}

\usepackage{acl} 

\usepackage{times}
\usepackage{latexsym}

\usepackage[T1]{fontenc}

\usepackage[utf8]{inputenc}

\usepackage{microtype}

\usepackage{inconsolata}

\usepackage{listings}
\usepackage{expex}

\usepackage{csquotes}

\usepackage{placeins}

\newcommand*{\subscript}[1]{\ensuremath{_\textrm{{\scriptsize #1}}}}

\usepackage{graphicx}

\usepackage{booktabs}

\title{A Scalable Pipeline for Estimating Verb Frame Frequencies Using Large Language Models}

\author{Adam M. Morgan \\
  NYU Grossman School of Medicine \\ 550 1\textsuperscript{st} Ave, New York NY 10016 \\
  \texttt{Adam.Morgan@NYULangone.org} \\\And
  Adeen Flinker \\
  NYU Grossman School of Medicine \\ 550 1\textsuperscript{st} Ave, New York NY 10016 \\
  NYU Tandon School of Engineering \\ 6 MetroTech Center, Brooklyn NY 11201 \\
  \texttt{Adeen.Flinker@NYULangone.org} \\}

\begin{document}
\maketitle

\begin{abstract}

We present an automated pipeline for estimating Verb Frame Frequencies (VFFs), the frequency with which a verb appears in particular syntactic frames. VFFs provide a powerful window into syntax in both human and machine language systems, but existing tools for calculating them are limited in scale, accuracy, or accessibility. We use large language models (LLMs) to generate a corpus of sentences containing 476 English verbs. Next, by instructing an LLM to behave like an expert linguist, we had it analyze the syntactic structure of the sentences in this corpus. This pipeline outperforms two widely used syntactic parsers across multiple evaluation datasets. Furthermore, it requires far fewer resources than manual parsing (the gold-standard), thereby enabling rapid, scalable VFF estimation.  Using the LLM parser, we produce a new VFF database with broader verb coverage, finer-grained syntactic distinctions, and explicit estimates of the relative frequencies of structural alternates commonly studied in psycholinguistics. The pipeline is easily customizable and extensible to new verbs, syntactic frames, and even other languages. We present this work as a proof of concept for automated frame frequency estimation, and release 
\href{https://www.doi.org/10.17605/OSF.IO/FRQBE}{all code and data}  
to support future research.

\end{abstract}

\section{Introduction}

Word (or \textit{lexical}) frequency is one of the most widely-used constructs in natural language research. 
In NLP, explicit use of word frequency estimates has driven major improvements in language models, increasing model speed by orders of magnitude \cite{mikolov2011extensions, mikolov2013efficient} and bringing models closer to human-like performance benchmarks \cite{pennington2014glove, gong2018frage}.  
In research on human cognition, it is one of the strongest predictors of behavior (e.g., response times \cite{balota1984lexical, brysbaert2018word}, event-related potentials \cite{van1990interactions}).  Its use led to discoveries of phenomena like the frequency-by-regularity interaction \cite{seidenberg1989distributed}, which paved the way to the development of the connectionist architecture underlying nearly all modern models of human cognition \cite{smolensky1988proper}. 

Where lexical frequency has proved invaluable for understanding word-level information, Verb Frame Frequencies (VFFs) -- the frequency with which a verb takes particular sets of arguments like a direct or indirect object\footnote{We define verb frames (also referred to as \textit{argument structures}, \textit{subcategorization frames}, or \textit{complement structures}) as the \textit{selected} arguments of the verb -- i.e., excluding optional modifiers like adjuncts and non-selected adverbs/adjectives.}  -- offer a powerful window into syntax, which remains far less understood than words. 
For instance, in behavioral research (\citealp[e.g.,][]{trueswell1998prune, berkovitch2019subliminal}), VFFs have been used to demonstrate that verb representations are inextricably linked to the structures they co-occur with, supporting ``lexicalist'' theories of syntax \cite{macdonald1994lexical, levin1994unaccusativity, pickering1998representation, ryskin2017verb}, and in neuroscientific research (\citealp[e.g.,][]{meltzer2015brain, shetreet2007cortical}) they have been leveraged to map syntactic and semantic information in the brain. 

However, relatively few studies have leveraged VFFs, and syntax remains far less understood than words. 
The main limiting factor is the limited availability of high-quality VFF estimates. That is, while there exists a vast number of tools and databases for calculating words' frequencies \citep[e.g.,][]{brysbaert2009moving, baayen1996celex, davies2008corpus, michel2011quantitative, balota2007english, coltheart1981mrc}, there are very few resources for obtaining good estimates of VFFs.  
Some resources provide detailed inventories of verb-frame types and fine-grained semantic distinctions (e.g., FrameNet \cite{ruppenhofer2016framenet}, VerbNet \cite{schuler2005verbnet}, VerbAtlas \cite{di2019verbatlas}), but they do not quantify how frequently individual verbs appear in each frame, a critical distinction in settings where frequency modulates cognitive processing (e.g., syntactic priming, disambiguation, language acquisition). 
Previous studies using VFFs have largely run their own norming tasks to obtain estimates (\citealp[e.g.,][]{trueswell1998prune, garnsey1997contributions, ryskin2017verb}), requiring significant time and effort.  
And while many such studies have made the raw data publicly available, these datasets are often small, involving just the subset of verbs and frames relevant to the particular study. 
Despite major advances in large-scale lexical and semantic resources, no existing tool provides scalable, empirically-grounded estimates of verb frame frequencies.

The primary obstacle to obtaining high-quality VFF estimates is the difficulty of \textit{parsing}, or identifying the underlying syntactic structure in a string of words.  Manual parsing, where trained linguists analyze and annotate a corpus (creating a ``treebank'') is the gold standard, but it is prohibitively time- and resource- intensive.  A number of automated approaches exist (\citealp[e.g.,][]{petrov2007improved, qi2020stanza, kitaev2018constituency}), but these ``parsers'' still require manual checking \cite{taylor2003penn} and suffer from systematic biases, including high error rates for certain complex structures \cite{rimell2009unbounded, choi2015depends} and in different language registers (e.g., informal speech; \citealt{yang2015domain}).  Consequently, while automated parsing provides valuable scalability and efficiency, the gold standard for estimating VFFs remains manual annotation. 

In the most comprehensive set of manual VFFs to date, \citet{gahl2004verb} had four trained linguists parse 200 sentences for each of 281 American English verbs.  They focused in particular on verbs implicated in a widely-studied source of processing difficulty: the Noun Phrase (NP)/Sentential Complement (SC) ambiguity, where a verb like \textit{accept}, which can take a NP or SC complement, is immediately followed by a NP like ``the money,'' which is temporarily ambiguous between an NP complement or the subject of an SC:
\pex \textit{The NP/SC Ambiguity:} \label{ex:np_sc}
\a accept [the money]\subscript{NP} \hfill \textit{NP} 
\a accept [the money is gone]\subscript{SC} \hfill \textit{SC}
\xe
They then compared their results to those of 10 earlier studies, demonstrating overall high agreement.  These frequencies have proven of immense value, having been used in dozens of subsequent behavioral and neuroscientific investigations of language processing (\citealp[e.g.,][]{staub2006heavy, linzen2016uncertainty, vuong2015role}).  

However, there are some notable gaps in the \citet{gahl2004verb} dataset.  
For one, \citeauthor{gahl2004verb} collapsed across intransitive and Prepositional Phrase (PP) frames (e.g., \textit{``look at me''} or \textit{``look for the remote''}), reducing the accuracy of their dataset for two of the most common verb frames. 
Perhaps even more importantly, they did not include verbs or frames implicated in one of the most commonly-studied phenomena in psycholinguistics: \textit{structural alternations}, where the production system can express the same meaning with different verb frames.  For instance, dative verbs like \textit{loan} can take a ``Direct Object'' (DO; Ex. \ref{ex:do_po:do}) or ``Prepositional Object'' (PO; Ex. \ref{ex:do_po:po}) frame, and locative verbs like \textit{load} can appear with an ``On'' (Ex. \ref{ex:on_with:on}) or ``With'' frame (Ex. \ref{ex:on_with:with}).  Consequently, their dataset lacks many common verbs (e.g., \textit{give}, \textit{put}, and \textit{show}) and frames that are among the most important in psycholinguistic research. 
\pex \textit{The Dative Alternation:} \label{ex:do_po}
\a loan [the kid]\subscript{NP} [a book]\subscript{NP} \hfill \textit{DO} \label{ex:do_po:do}
\a loan [a book]\subscript{NP} [to the kid]\subscript{PP} \hfill \textit{PO} \label{ex:do_po:po}
\xe 
\pex~ \textit{The Locative Alternation:} \label{ex:on_with}
\a load [hay]\subscript{NP} [onto the truck]\subscript{PP} \hfill \textit{``on''} \label{ex:on_with:on}
\a load [the truck]\subscript{NP} [with hay]\subscript{PP} \hfill \textit{``with''} \label{ex:on_with:with}
\xe

Critical to understanding how producers choose between competing structures is knowing the relative frequency of the competing alternates.  However, few resources exist for these alternating frames. 
Some studies have attempted to circumvent the difficulty of parsing large datasets.  For instance, \citet{hawkins2020investigating} used LLM surprisal values to assess the relative preference of English dative verbs for the Direct Object (DO) and Prepositional Object (PO) frames (\ref{ex:do_po}). 
To evaluate performance, rather than running a production study or parsing a corpus, \citet{hawkins2020investigating} asked human participants to use a slider to indicate their relative preference for a DO sentence vs. its corresponding PO formulation.  
In not requiring transcription and parsing, this approach enables the rapid collection and analysis of a huge amount of data.  Their results showed strong correlations between the human preference ratings and relative surprisals for a number of the language models, explaining 73\% of the variance with GPT2-large.  
However, questions remain about the interpretation of relative preference ratings, including their degree of psychometric validity and how exactly they relate to frequency (\citealp[cf.][]{myers2017acceptability, white2020frequency}). 

In sum, there are three main limitations in extant tools for estimating VFFs.  First is the trade-off between high quality parses and scalability: manually annotated datasets are the gold-standard for calculating VFFs, but being extremely time- and labor-intensive, these cannot readily be adapted or extended to, e.g., additional verbs, frames, or other languages.  Second, existing datasets have important gaps, either in the granularity of their syntactic distinctions (as in \citeposs{gahl2004verb} collapsing of intransitive and PP frames) or in the breadth of their coverage (e.g., the absence of dative and locative verbs and frames).  Third, existing studies have used vastly different methodological approaches (e.g., manual annotation, relative acceptability rating, LLM surprisals), making it difficult to evaluate findings across studies or combine results to build more comprehensive datasets. 

Here we aim to overcome these limitations by leveraging recent advances in artificial intelligence to create an automated pipeline for calculating VFFs that is fast, accurate, customizable, and scalable. We began by compiling a more comprehensive list of verbs, including all 281 verbs in \citeauthor{gahl2004verb} as well as 195 additional verbs implicated in the dative and locative alternations.  We then automated the creation of a mini-corpus, using an LLM to repeatedly generate sentences given the entire set of 476 verbs. By instructing the LLM to behave like an expert linguist, we obtained parses for each sentence, converting our corpus into a treebank. To evaluate how well this pipeline performed relative to existing tools, we parsed the same sentences with two commonly-used constituency parsers: the Berkeley Neural Parser (\textit{benepar}; \citealt{kitaev2018constituency, kitaev2019multilingual}) and Stanford CoreNLP \cite{manning2014stanford}.  We then compare the results from the LLM and the two existing parsers to previously published datasets and show that in nearly every case, the LLM significantly outperforms the Berkeley and Stanford Parsers.  Alongside this manuscript, we make available all scripts used in this pipeline, as well as the raw, preprocessed, and categorized parses. By making VFF estimation fast and scalable, our approach enables future researchers to choose their own level of granularity among syntactic distinctions, and to readily scale up to other verbs, frames, and even other languages. 

\section{Methods}

\subsection{Data Sources}

\paragraph{Verb selection.}
We selected 476 English verbs from established lexical databases frequently used in linguistic and psycholinguistic research.  This included all 281 verbs used in \citeposs{gahl2004verb} gold-standard dataset of American English verb frame frequencies, as well as verbs included in previous studies of the dative alternation \cite{hawkins2020investigating, theijssen2009regression}, locative alternation, and NP-X ambiguity \cite{trueswell1993verb, garnsey1997contributions}, as well as those associated with these alternations in \citealp{levin1993english}. These sources were chosen for their broad acceptance and foundational role in verb argument structure and syntactic processing research.

\paragraph{Context Generation.}
We used Open AI's GPT-o1 to generate 1,000 brief task contexts (see Appendix \ref{sec:appendix:context_generation} for the complete prompt).  The output included contexts describing various locations, situations, and times, including, e.g., ``on the tennis court'' and ``down to the wire.''

\paragraph{Sentence Generation via LLM.}
Sentences using each verb were created using Open AI's GPT-4o-mini \cite{openai2024gpt4osystemcard}. The system message instructed the model: ``You are a random sentence generator, tasked with generating natural-sounding sentences like those you might find in a conversation, movie, or newspaper. The random sentence generator should provide sentences with varied meanings, tenses, and syntactic structures, simulating random draws from sentences anywhere on the internet -- including in movies, newspapers, conversations, forums, etc. to produce naturalistic, plausible sentences using a specific verb and context.''  It further specified, ``Responses do not have to include words from the context, but please ensure they are thematically related,'' and listed four example input-output pairs (see Appendix \ref{sec:appendix:sentence_generation} for the full system message).  A total of 100 batch prompts were sent via the API, each of which included the full list of 476 verbs along with a single context, generating 100 sentences for each verb.

\paragraph{Risk statement.}
This project poses minimal risk. However, because it generates and parses sentences, biases in the LLM’s training data likely leads to over- or underrepresentation of dialect-specific verb biases. By releasing all code and data, we aim to promote transparency. Future work should assess and mitigate such biases.

\subsection{Parsing}

\paragraph{Sentence Preprocessing.} 
Prior to parsing, sentence strings were split by commas, which typically indicated root-level clause boundaries. The resulting substrings that did not include the main verb were discarded.  

\paragraph{The Berkeley Neural Parser.} 
The clause containing the target verb was parsed using the Berkeley Neural Parser (benepar; \citealp{kitaev2018constituency}), implemented via the spaCy interface with the benepar\_en3 model (see Script 3b in the repository). Benepar assigns Penn Treebank–style constituency structures to text using a neural chart parser trained on annotated corpora. It maps words to hierarchical phrase structures annotated with clause- and phrase-level labels from the Penn Treebank tagset \citep{taylor2003penn}.

\paragraph{The Stanford CoreNLP Constituency Parser.} 
The clause with the target verb was also passed to the Stanford CoreNLP constituency parser (version 4.5.7) using the NLTK interface (version 4.2.0; \citealp{bird2006nltk}) (see Script 1 in the repository). Like benepar, the Stanford Parser maps words to hierarchical structures, with nodes labeled using the Penn Treebank tags. 

\paragraph{The LLM (GPT-4o) Parser.} 
Argument structures were also annotated using GPT-4o (specifically: gpt-4o-2024-05-13; \citealp{openai2024gpt4osystemcard}). Initial attempts at feeding all sentences for each of the 100 sentence generation batches to the model in a single batch resulted in the model omitting a large proportion of the input sentences in its response.  Trial and error revealed that the model was more likely to return a parse for each input sentence when the number of sentences was 100 or fewer, so we split the prompts into sets of 100 or fewer sentences.

The system message (which appears verbatim in Appendix \ref{sec:appendix:llm_parse}) instructed GPT-4o that it was an expert linguist who would be given multiple sentence-verb pairs.  For each entry, its task was to isolate the clause containing the given verb, remove everything outside of the Verb Phrase (VP) as well as any optional modifiers like ``time, manner, or location expressions that are not required or licensed by the verb,'' and to return just the arguments that the verb subcategorizes for.  We provided several examples, which included a diversity of argument structures as well as situations where optional arguments (e.g., adjectives) should be ignored.  
The model was instructed to tag each argument using the standard Penn Treebank tags to facilitate comparison to the output of the Stanford and Berkeley Parsers. 
We explicitly instructed the model not to infer missing arguments to prevent the model from annotating a sentence like ``I ate'' as transitive (i.e., having an object like ``dinner''), given that, even though there is an implicit semantic object, syntactically this is an intransitive use of the verb \textit{eat}.  The full system message is provided in Appendix \ref{sec:appendix:llm_parse}.

\subsection{Data Cleaning} \label{sec:methods:cleaning}

Prior to categorizing argument structures, we cleaned the responses from both the Stanford, Berkeley, and GPT-4o parsers (Script 4). 
To avoid potential hallucination of arguments in the LLM parses, we explicitly verified that every argument the LLM identified was present in the original sentence. 
We excluded sentences where the target verb was incorrectly used as a noun (e.g., \textit{party}; ``I never thought a party could turn so somber.''). 
Prepositional phrases were labeled by their particular preposition to distinguish between ordinary prepositional arguments (e.g., ``cook [carrots]\subscript{NP} [on the stove]\subscript{PP}'') and those that are implicated in the alternations of interest (e.g., \textit{for}-phrases in benefactive constructions like ``cook [carrots]\subscript{NP} [for the children]\subscript{PP}'' -- which could alternatively appear in the Double Object construction: ``cook [the children]\subscript{NP} [carrots]\subscript{NP}). 
In all, these steps removed 3,087 sentences from the dataset (4.779\% of the original 64,589). 

Finally, we excluded 75 sentences where the target verb appeared in a passive construction. This decision was motivated by two concerns. First, both parsers showed inconsistent behavior in annotating passives, leading to unreliability in frame assignment. Second, there is no clear consensus on how passives should be categorized: while syntactically intransitive, many linguistic theories (e.g., Chomsky, 2014a,b) argue that they are derived from underlying transitive structures. Given these inconsistencies and theoretical ambiguities, and because passives represented a small proportion of the data, we excluded them to improve comparability and reliability of frame estimates.

For verbs implicated in the Dative or Locative alternations, we separately estimated their rates of each alternating structure.  This involved certain additional criteria.  As the PO dative alternation (Ex \ref{ex:do_po:po}) only involves PPs headed by \textit{to} or \textit{for}, we excluded PPs headed by other prepositions.  Similarly, the locative alternation only involves a subset of prepositions for each verb (e.g., \textit{with}, \textit{on(to)}, \textit{in(to)}, \textit{over}, etc.). We excluded NP-PP frames where the PP was headed by a preposition that the specific verb does not take in locative constructions.

\subsection{Argument Structure Extraction and Frame Categorization} \label{sec:methods:categorization}

To calculate VFFs, we counted the occurrences of each of the unique argument structures across the cleaned dataset.  To simplify matters, we established a minimum threshold for the number of times each argument structure had to appear in the dataset to be counted as a unique frame (specifically, the 75\textsuperscript{th} percentile of the frame counts, or 65 instances).  We binned all frames that did not meet this threshold into an ``other'' category.

\subsection{Evaluation and Comparison to Existing Norms}

Our automated pipeline's output was validated against the gold-standard Verb Frame Frequencies reported in \citet{gahl2004verb}. \citeauthor{gahl2004verb}'s dataset consisted of manually parsed sentences (200 sentences for each of 281 verbs), providing a reliable benchmark for assessing our method’s accuracy. Comparative analyses included quantitative evaluations of frame frequency distributions and qualitative examinations of discrepancies attributable to methodological differences. 

Our method diverged from \citeposs{gahl2004verb} in two important ways.  First, due to the importance of structural alternations in the psycholinguistic literature, we added a number of verbs to our dataset implicated in either the locative alternation (verbs like \textit{stuff} and \textit{spray}) or the dative alternation (including ``transfer'' verbs like \textit{give} and \textit{hand} and ``benefactive'' verbs like \textit{bake} and \textit{buy}).   
Second, whereas \citeauthor{gahl2004verb} treated verbs which select for Prepositional Phrases (PPs; e.g., ``look [for my keys]\subscript{PP}'') as intransitive, here we treat these as their own subcategorization frame, distinct from intransitive and NP frames. For the purposes of evaluating our pipeline's output, we relabeled all PP frames as intransitive prior to comparison to \citeauthor{gahl2004verb}'s data.

\section{Results}

All data, scripts, and results are available online at 
\url{https://www.doi.org/10.17605/OSF.IO/FRQBE}. 

\subsection{Unique Subcategorization Frames}

\begin{figure}[ht!]
\centering
\includegraphics[width=0.36\textwidth]{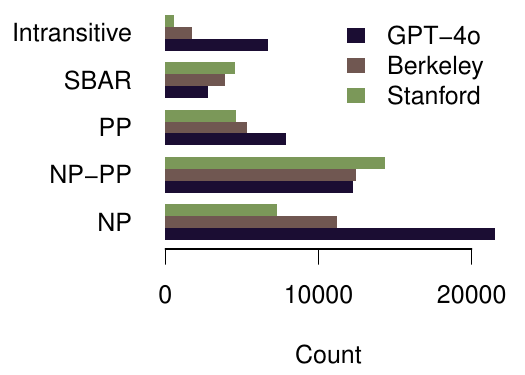}
\caption{Counts of the five most frequent verb frames (out of 61,427 sentences), by parser.}
\label{fig:frequent_frames}
\end{figure}

After cleaning the data (Section \ref{sec:methods:cleaning}) and binning low-frequency frames into an ``other'' category (Section \ref{sec:methods:categorization}), we were left with 81 unique frames.  Figure \ref{fig:frequent_frames} shows the number of instances of the 5 most common frames. Sample sentences for each of these appear in Appendix \ref{sec:appendix:frequent_frames}. 
Notably, two of these frames involve PP arguments, which \citet{gahl2004verb} excluded from their frame counts. By contrast, our approach treats selected PPs as syntactically licensed arguments, resulting in a more fine-grained and psycholinguistically relevant set of verb frame categories (see Section \ref{sec:results:intransitivity}). 

\subsection{Evaluating the Parsers}

Despite being applied to the same set of sentences, GPT-4o produced the same parse as the the Berkeley Parser for only 45\% of sentences, and only 37\% for the Stanford Parser. To evaluate accuracy, we compared results from all three parsers to the manually parsed (i.e., gold-standard) \citet{gahl2004verb} dataset. 
(The Berkeley Neural Parser being state-of-the-art, we report these comparisons in the main text and provide Stanford results in the supplement for comparison.) 
Figure \ref{fig:scatter:NP} shows the relationship between the proportion of NP (i.e., transitive) tags between each verb in \citeauthor{gahl2004verb} and the GPT-4o (left) and Berkeley (right) parsers. 
We found significant relationships for both GPT-4o (linear regression, $t(245) = 13.225$, $p<.001$) and Berkeley (linear regression, $t(245)=7.502$, $p<.001$), although the GPT-4o VFFs accounted for over twice as much variance ($r^2 = .417$) as the Berkeley VFFs ($r^2 = .187$). 

\begin{figure}[ht!]
\centering
\includegraphics[width=0.22\textwidth]{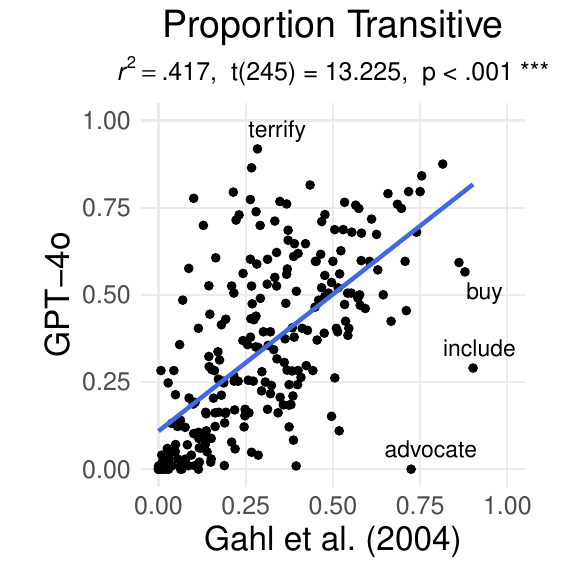}
%
\includegraphics[width=0.22\textwidth]{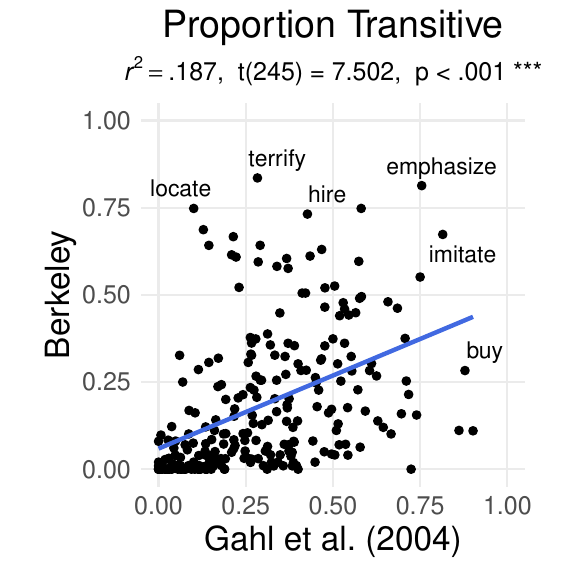}
\caption{Evaluating the GPT-4o and Berkeley Parsers' VFF estimates.  We compared estimates to \citeposs{gahl2004verb} gold-standard dataset.  Here we show these comparisons for the NP frame (see Appendix \ref{sec:appendix:scatterplots} for others).  Results showed significant correlations for both parsers, but GPT-4o (left) produced a better fit ($r^2=.406$) than the Berkeley Neural Parser (right; $r^2=.187$).}
\label{fig:scatter:NP}
\end{figure}

We repeated this analysis for each of the seven modal frames in Gahl et al.'s dataset (see Appendix \ref{sec:appendix:scatterplots}).  
This resulted in a total of 14 ``univariate'' models: seven structures $\times$ two parsers.  Our results showed that both parsers performed above chance: the GPT-4o VFFs significantly predicted \citeposs{gahl2004verb} for all seven frames, and the Berkeley VFFs were significant for six frames (see Table \ref{tab:univariate_gahl_results} in Appendix \ref{sec:appendix:scatterplots}).  However, relative to the Berkeley estimates, GPT-4o estimates accounted for more variance in the \citet{gahl2004verb} data for every structure we tested -- on average,   
1.454 times more, and 2.010 times more than the Stanford estimates.

To statistically compare how well the GPT-4o and Berkeley Parsers fit \citeposs{gahl2004verb} data, we performed a series of model comparisons.  
We started by re-fitting the linear models using Bayesian regression \cite{burkner2017brms} and, for each verb frame, calculated the Bayes Factor (BF), or how many times more evidence there was for the GPT-4o VFFs than the Berkeley VFFs. 
For all verb frames, there was ``strong'' or ``decisive'' evidence ($log_{10}$(BF) $>1$ and $>2$, respectively; \citealt{jeffreys1998theory}) in favor of the GPT-4o VFFs over the Berkeley VFFs. 
Taken together, these results suggest that GPT-4o, when prompted appropriately, can outperform the most widely used automated parsers.

\begin{table}[ht!]
\centering
\setlength{\tabcolsep}{3pt} 
\begin{tabular}{lcrl}
\toprule
\textbf{}    
                & \textbf{Winning}    
                            & \textbf{} 
                                            & \textbf{Degree of} \\
\textbf{Structure}    
                & \textbf{Parser}    
                            & \textbf{\textit{log}\subscript{10}(BF)} 
                                            & \textbf{Evidence} \\
\midrule  
Transitive      &  GPT-4o   & 17.490        & Decisive \\
Intransitive    &  GPT-4o   & 29.521        & Decisive \\
Sentence        &  GPT-4o   & 1.328         & Strong \\
Particle        &  GPT-4o   & 3.301         & Decisive \\
NP-Sentence     &  GPT-4o   & 7.351         & Decisive \\
Non-finite      &  GPT-4o   & 3.454         & Decisive \\
Particle-NP     &  GPT-4o   & 5.716         & Decisive \\
\bottomrule
\end{tabular}
\caption{Model comparisons. For seven structures, we compared models predicting \citeposs{gahl2004verb} gold-standard VFFs using the GPT-4o or Berkeley VFFs.  The resulting Bayes Factors (BFs) indicate how much more evidence there was for the GPT-4o VFFs than the Berkeley VFFs (positive values in log-space) or vice versa (negative).  BFs were interpreted according to the \citet{jeffreys1998theory} scale.  For six of the seven structures, we found ``decisive'' evidence that the GPT-4o estimates fit \citeauthor{gahl2004verb}'s VFFs better than the Berkeley estimates. For the sentence frame, evidence was merely ``strong.''}
\label{tab:model-comparison}
\end{table}

\subsection{Human Validation}

\begin{figure}[ht!]
\centering
\includegraphics[width=0.34\textwidth]{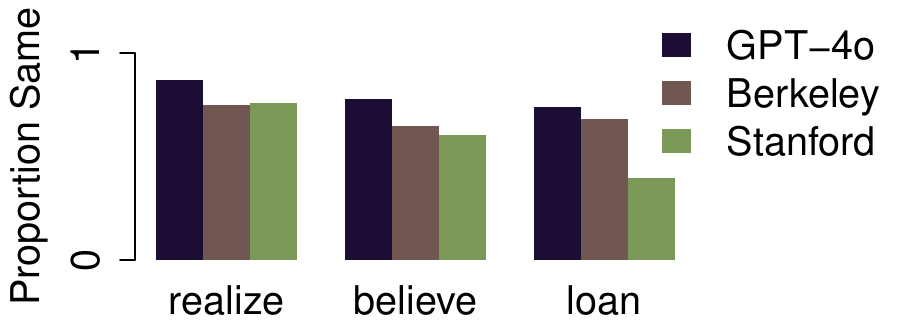}
\caption{Human validation: The proportion of sentences for which each parser agreed with human annotations for three representative verbs (100 sentences per verb). GPT-4o consistently outperformed the other two.}
\label{fig:manual}
\end{figure}

To evaluate parser accuracy against human annotations, an expert linguist manually annotated 300 sentences -- 100 for each of three representative verbs: \textit{realize}, \textit{believe}, and \textit{loan}. We then compared these manual annotations to the frame predictions made by each of the three automated parsers. GPT-4o achieved the highest match rate with manual annotations for all three verbs (Fig. \ref{fig:manual}), agreeing on 79\% of sentences overall, compared to 69\% for Berkeley and 59\% for Stanford. Critically, GPT-4o agreed with the human annotations on 74\% of sentences for the verb \textit{loan}, which is not present in \citeauthor{gahl2004verb}'s dataset, demonstrating generalizability to a dative verb (Ex. \ref{ex:do_po}) commonly used in psycholinguistic experiments.

\subsection{The NP/SC Ambiguity}

\begin{figure}[ht!]
\centering
\includegraphics[width=0.22\textwidth]{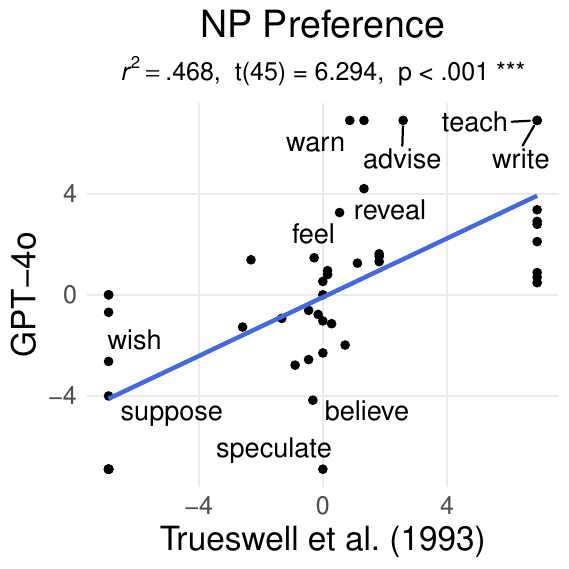}
\includegraphics[width=0.22\textwidth]{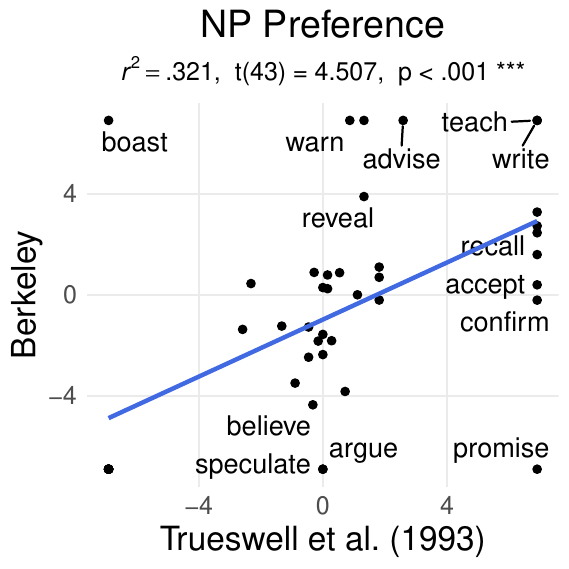}
\vspace{5pt}

\includegraphics[width=0.22\textwidth]{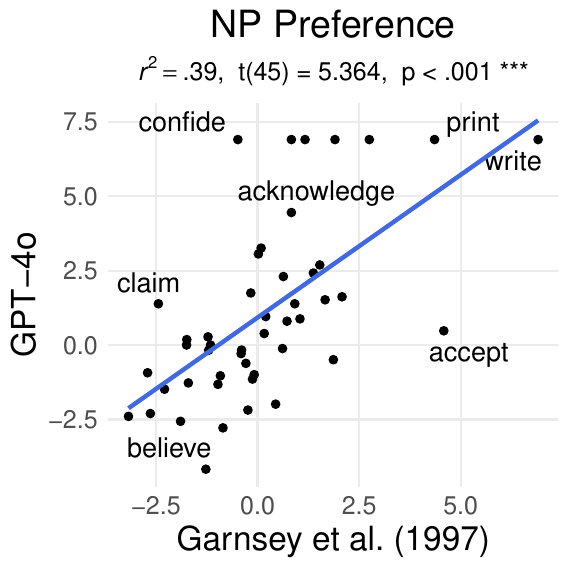}
\includegraphics[width=0.22\textwidth]{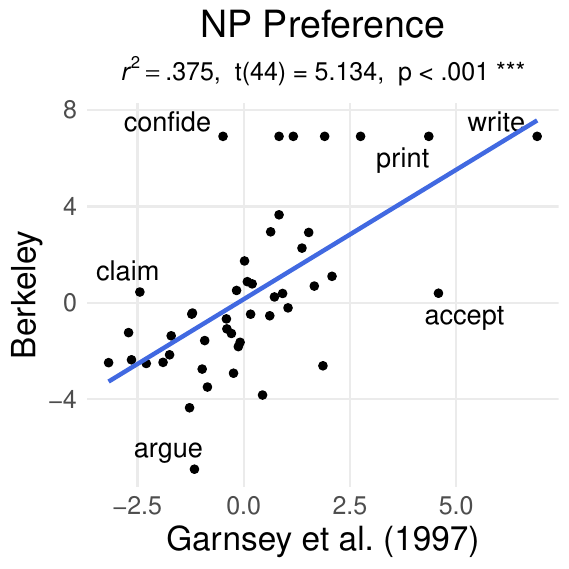}
\caption{Evaluating GPT-4o (left column) and Berkeley Neural Parser (right) estimates of NP bias for NP/SC verbs.  We compared our estimates to those from two prior studies: \citeauthor{trueswell1993verb} (\citeyear{trueswell1993verb}; top row) and \citeauthor{garnsey1997contributions} (\citeyear{garnsey1997contributions}; bottom). Both parsers significantly predicted the previous results, though the GPT-4o model accounted for more of the variance in both datasets.}
\label{fig:scatter:np_sc_ambiguity}
\end{figure}

Given the importance of the NP/SC ambiguity in psycholinguistic research, we sought to validate our estimates of verb biases for these two frames.  We started by excluding all responses that were not categorized as either an NP or SC completion for each parser, and then calculating the log-odds of NP over SC completions for each of $\sim 45$ verbs used in two prior studies that published norming data.  Figure \ref{fig:scatter:np_sc_ambiguity} shows that both the GPT-4o- and Berkeley-based estimates significantly predicted the results from \citet{trueswell1993verb} (GPT-4o: $r^2=.468$, $p<.001$; Berkeley: $r^2=.321$, $p<.001$) and \citet{garnsey1997contributions} (GPT-4o: $r^2=.390$, $p<.001$; Berkeley: $r^2=.375$, $p<.001$). 
Bayesian model comparison revealed ``decisive'' evidence in favor of the GPT-4o estimates when modeling \citeauthor{trueswell1993verb}'s data ($log_{10}$(BF)$=2.386$; \citealt{jeffreys1998theory}), but no noteworthy evidence in favor of either parser for modeling \citeauthor{garnsey1997contributions}'s data ($log_{10}$(BF)$=.290$). 
Overall, while both parsers capture verbs’ NP/SC biases to some extent, GPT-4o’s estimates more closely align with those obtained from these human norming studies, reinforcing its potential as a tool for accurately estimating verb frame frequencies.

\subsection{Improving Estimates of Intransitivity} \label{sec:results:intransitivity}

\begin{figure}[ht!]
\centering
\includegraphics[width=0.22\textwidth]{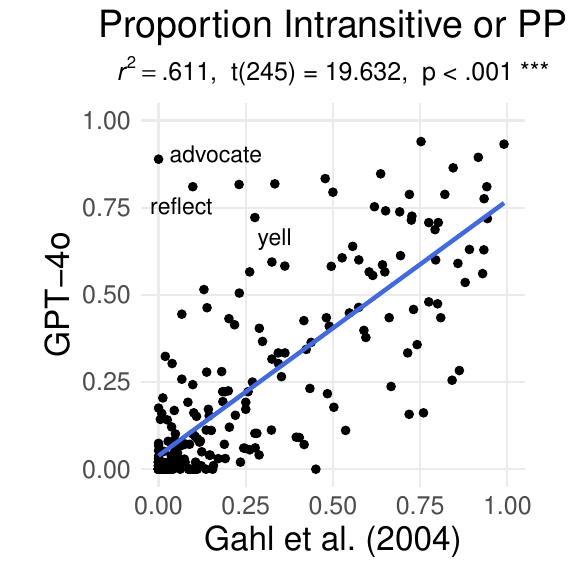}
\includegraphics[width=0.22\textwidth]{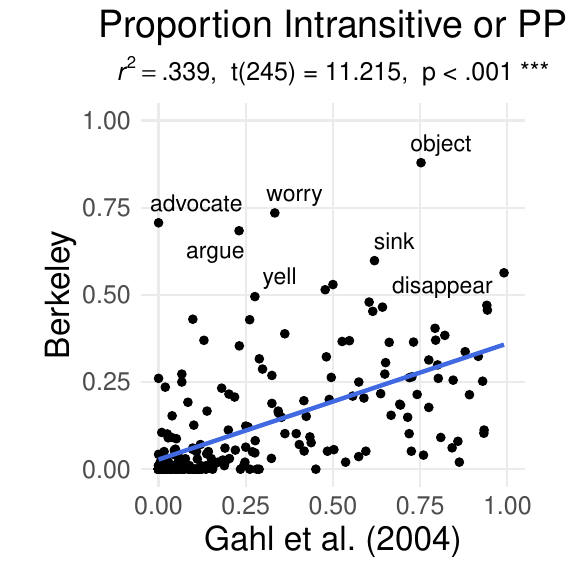}
\vspace{5pt}

\includegraphics[width=0.22\textwidth]{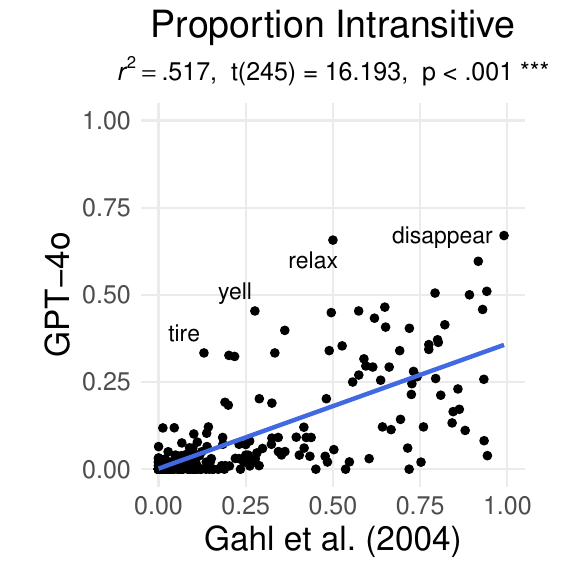}
\includegraphics[width=0.22\textwidth]{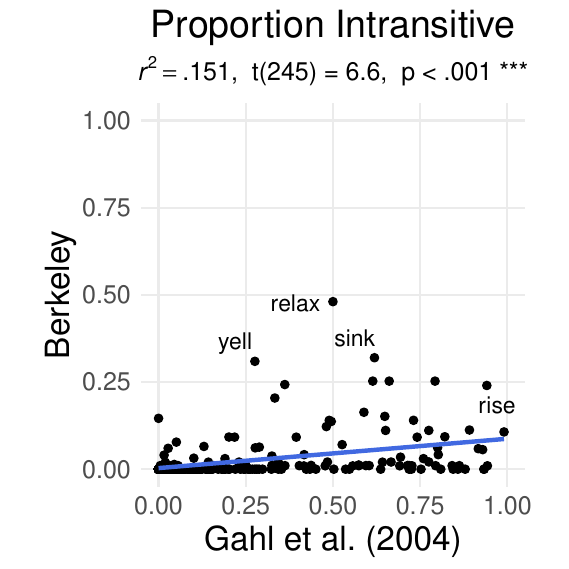}
\caption{Evaluating the effect of including PP frames in intransitivity counts.  For both the GPT-4o (left) and Berkeley (right) parsers, including PPs (as \citet{gahl2004verb} did) produced a better fit (top) to \citeauthor{gahl2004verb}'s estimates than excluding them (bottom), meaning that \citeauthor{gahl2004verb} lost meaningful information by collapsing these categories.}
\label{fig:granularity}
\end{figure}

One shortcoming of \citeposs{gahl2004verb} estimates is that they collapsed intransitive and PP frames.  Beyond reducing syntactic granularity, this may have led to systematic errors in their estimates of intransitivity.  To assess the impact on their results, we compared our intransitivity estimates to theirs in two ways: first, counting PP frames as intransitive (Fig. \ref{fig:granularity}, top row), and next counting only verbs with no arguments as intransitive (bottom).  As expected, excluding PPs from intransitivity counts produced worse fits for both parsers, reducing $r^2$ from .611 to .517 for the GPT-4o parser (left column) and from .339 to .151 for the Berkeley Neural Parser (right). 
These results suggest that separating PP frames improves the accuracy of intransitivity estimates, and that our method may provide more faithful estimates of intransitivity rates. 

\subsection{The Dative Alternation}

Another primary goal of this work was to provide high-quality VFF estimates for commonly studied verb frames that were not included in \citet{gahl2004verb} -- in particular, datives and locatives, which feature prominently in psycholinguistic research.  
We therefore added a number of verbs not present in the \citeauthor{gahl2004verb} dataset. 
We focused on two ditransitive frames that were not tracked in \citeauthor{gahl2004verb}: NP-NP (as in the Dative Direct Object frame; Ex. \ref{ex:do_po:do}) and NP-PP (as in the Prepositional Object frame; Ex. \ref{ex:do_po:po}).

To estimate verb-specific preferences for the two dative frames, we calculated the log-odds of DO over PO rates for the GPT-4o and Berkeley parses. Our results showed a wide range of preferences, with some verbs strongly preferring the DO frame (e.g., \textit{teach} as in ``teach the dog a trick'') and others preferring PO completions (\textit{ship} as in ``ship the crate to its owner''). 

\begin{figure}[ht!]
\centering
\includegraphics[width=0.22\textwidth]{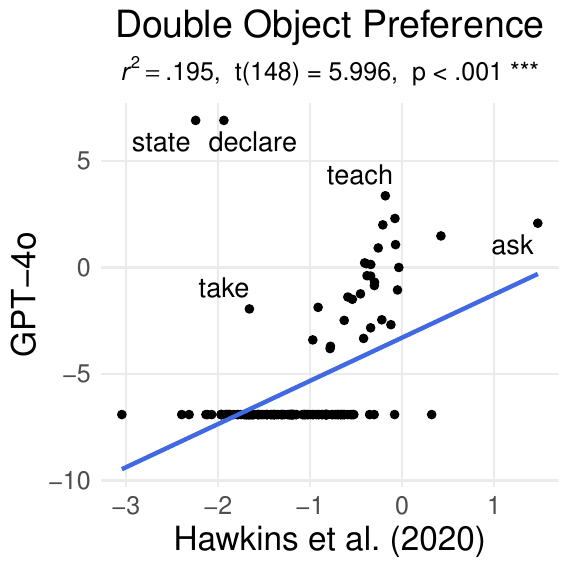}
%
\includegraphics[width=0.22\textwidth]{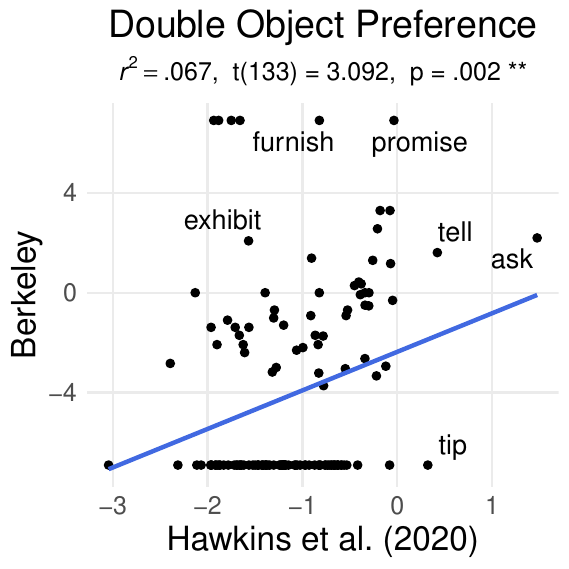}
\caption{Evaluating the GPT-4o (left) and Berkeley (right) Parsers' estimates of DO/PD bias for dative verbs.  We compared estimates to \citeposs{hawkins2020investigating} relative preference ratings.  Many of the verbs they included are non-alternating (e.g., \textit{whisper}), and for the majority of these neither GPT-4o nor the Berkeley parser identified any DO completions.  To avoid infinities, we set proportions of 0 (no instances) or 1 to .001 and .999, respectively, capping log-odds values at $\pm6.907$ (the row of black dots at the bottom of each plot).  Both parsers significantly predicted preference ratings, but GPT-4o accounted for more variance ($r^2=.215$) than the Berkeley Neural Parser ($r^2=.067$).}
\label{fig:scatter:datives}
\end{figure}

To evaluate the accuracy of these estimates, we compared them to the relative preference ratings reported in \citet{hawkins2020investigating} for the 150 verbs our studies shared in common. 
Many verbs had no DO completions at all (the row of dots at the bottom of both panels in Fig. \ref{fig:scatter:datives}), consistent with \citeauthor{hawkins2020investigating}’s inclusion of non-alternating PO verbs. 
As before, we performed a Bayesian model comparison, which revealed ``decisive'' evidence ($log_{10}$(BF)$=8.700$; \citealt{jeffreys1998theory}) in favor of GPT-4o over Berkeley.

\section{Discussion}

Verb Frame Frequencies (VFFs) have proven central to understanding the cognitive and neural underpinnings of syntax, yet they remain underutilized—especially when compared to lexical frequencies.   In machine language models, where they are nearly entirely unexplored, their explicit modeling stands to drastically improve model efficiency and performance as has been demonstrated for lexical frequency \cite{mikolov2011extensions, mikolov2013efficient, pennington2014glove, gong2018frage}.  The primary obstacle to such research is the difficulty of deriving high-quality estimates of VFFs.  
Manual annotation remains the gold standard, but is prohibitively time- and labor-intensive. 
Automated parsers offer a scalable alternative, but often exhibit systematic biases, particularly for rare or structurally complex constructions \cite{rimell2009unbounded, choi2015depends, yang2015domain}.  What such datasets exist are limited in various ways: syntactic granularity \cite{gahl2004verb}, interpretability of the metrics \cite{hawkins2020investigating}, or tailored to particular theoretical questions \cite{trueswell1993verb, garnsey1997contributions}, curbing the degree to which they can be extended for more general use.

In this work, we introduced a fully automated pipeline for estimating VFFs, leveraging recent advances in artificial intelligence for both sentence generation and syntactic parsing. By prompting an LLM to generate diverse, contextually grounded sentences using a target verb, and then instructing it to parse those sentences using linguistic conventions, we created a syntactically annotated corpus of over 45,000 sentences spanning 476 English verbs. This dual-use of the LLM represents a novel contribution, enabling scalable VFF estimation without any manual intervention.

We benchmarked this approach against three major sources of VFF norms: \citet{gahl2004verb} for broad coverage of argument types; \citet{trueswell1993verb} and \citet{garnsey1997contributions} for NP/SC ambiguities; and \citet{hawkins2020investigating} for datives. The LLM parser significantly predicted human data in all three cases, and in nearly every comparison outperformed the Berkeley Neural Parser (and the Stanford Parser; see Supplementary Section \ref{sec:appendix:scatterplots:stanford}). These results validate both steps of the pipeline: the use of LLMs to simulate naturalistic syntax in generation, and their ability to produce linguistically coherent parses. The pipeline’s ability to recover known alternation patterns across diverse verbs highlights its utility for psycho- and neuro-linguistic research, where fine-grained verb frame frequencies are critical but difficult to obtain at scale.

More broadly, this work lays the foundation for future research using VFFs in both cognitive science and NLP. The resulting database includes more verbs, finer-grained distinctions, and broader coverage of structural alternations than any existing dataset. It can be readily extended to new verbs, languages, or syntactic phenomena. As tools like LLMs continue to improve, this pipeline could enable rapid, domain-specific estimation of syntactic preferences—supporting applications from psycholinguistic modeling to improved syntactic generalization in neural architectures.


\section*{Limitations} 

This study constitutes a first attempt at building a verb-frame frequency database using large language models (LLMs) for both corpus generation and syntactic parsing.    
By automating both steps, the pipeline offers a scalable alternative to the previously labor-intensive processes of dataset creation, curation, and manual parsing. However, because the same model is used for generation and parsing, it is difficult to disentangle whether any observed limitations in accuracy stem from one stage or the other. Future work might separate these steps -- e.g., by parsing human-written corpora or applying different models at each stage—to better isolate their contributions.

Our approach also involves a number of assumptions that merit further evaluation.  To automate corpus creation for a targeted set of verbs, we used an LLM to generate sentences given those verbs. This assumes that the syntactic distributions produced by the model approximate those found in natural language environments. While this assumption is plausible given the size and breadth of LLM training corpora, the degree of variability in LLM outputs is modulated by hyperparameters that we did not manipulate (e.g., temperature, top\_$p$, top\_$k$).  
Our results showed strong correlations with existing corpora, but future work could improve performance by tuning these parameters to better match naturalistic distributions.

Similarly, we assumed that the model can provide reliable syntactic parses when instructed to behave like a linguist. While the LLM-generated parses aligned well with gold-standard data, this does not preclude the possibility of systematic biases in how the model analyzes sentence structure. Identifying such biases would be an important step toward improving accuracy further.

A related limitation is that the pipeline includes many degrees of freedom, any of which could impact the results.  
This leaves open many avenues for future research, for instance, evaluating the impacts of model parameters like temperature, nucleus sampling, token sampling, max tokens, and the frequency penalty; different model architectures (e.g., decoder-only transformers like GPT-3 or LLaMa vs. encoder-decoder models like T5 or BART); and input-level decisions (e.g., verb form, context inclusion, prompt phrasing, or the number of examples per request). Here we focused on introducing the method and demonstrating its potential for achieving high quality results with minimal manual effort. Future work should systematically explore these design variables.

An anonymous reviewer points out that, since both sentence generation and frame extraction are performed using models from the GPT family, it is possible that syntactic patterns in the generated data may align closely with the model’s own parsing expectations, potentially inflating parsing success. To reduce potential circularity, we used different models for each stage: GPT-4o-mini for generation and GPT-4o for parsing. While both belong to the same family, they differ in architecture, capacity, and likely training exposure, reducing the chance that parsing simply reflects generation structure. Still, it is likely there is some degree of representational overlap. Future work will test this directly using models from entirely different families (e.g., LLaMA 3) for parsing.

One further limitation with respect to \citet{gahl2004verb} is that we did not include passives.  It remains unclear exactly how to determine which syntactic frame a passive verb belongs to: for instance, is ``the melon was eaten'' transitive, since the canonical object of ``eat'' is ``melon''?  Or is ``eat'' intransitive since ``melon'' is the syntactic subject here?  And what is the relationship between each of these possibilities and the neural and model implementation of VFFs?  In response to these unknowns, previous work like \citeauthor{gahl2004verb} has separately estimated VFFs for each possibility.  In our case, both parsers showed inconsistent treatment of passives, further complicating analysis. For simplicity and clarity, we excluded them from the present study. 

Finally, our parser comparison focused on a single baseline: the Stanford CoreNLP constituency parser. However, many widely used parsers exist, including spaCy \cite{spacy2020}, Stanza \cite{qi2020stanza}, AllenNLP, and Berkeley’s neural parser \cite{kitaev2018constituency}. 
Each offers distinct strengths and may outperform the Stanford parser on certain constructions. Future research should benchmark LLM-based parsing against these alternatives, as it may be the case that even higher accuracy can be achieved from hybrid pipelines: for instance, using an LLM to generate a corpus given verbs/criteria of interest, and then using an existing automated parser to parse the output.


\section*{Ethics Statement} 

We take seriously ethical considerations in research and computing.  The present work strictly abides by the \href{https://www.aclweb.org/portal/content/acl-code-ethics}{ACL Ethics Policy}.

\section*{Acknowledgements} 

We thank Andrew Chang and Jean Pouget-Abadie for invaluable guidance on this project. 

\bibliography{treebank_v4}

\appendix

\section{Context generation}\label{sec:appendix:context_generation}

For context generation, we used Open AI's ChatGPT interface to the GPT-o1 model \cite{openai2024o1systemcard} on January 25th, 2025 with the following prompt:

\textit{
I'm working on a norming task where I ask participants to produce words in sentences. I want to provide participants with a short context -- just a simple phrase like ``on the tennis court,'' ``debating vaccine mandates,'' or ``about to give birth.'' I will instruct them to say the first sentence that comes to mind using given a word and one of these contexts. Please help me come up with contexts by generating a CSV with two columns: ``number'' and ``context''. The ``number'' column should number the contexts from 1 to 1000, and the ``context'' column should contain 1000 different contexts, all roughly 3-4 words long. After generating this CSV, please go back and ensure that there are no duplicates.
}
    
\section{Sentence generation}\label{sec:appendix:sentence_generation}

\subsection{System message}

\begin{lstlisting}[breaklines]
You are a random sentence generator, tasked with generating natural-sounding sentences like those you might find in a conversation, movie, or newspaper.
The random sentence generator should provide sentences with varied meanings, tenses, and syntactic structures, simulating random draws from sentences anywhere on the internet -- including in movies, newspapers, conversations, forums, etc.
I will provide you with a short context and a numbered list of exactly " + str(n_verbs_expected) + " verbs.
You must return exactly " + str(n_verbs_expected) + " lines, one line per verb in the same order, each of which uses the verb of the corresponding number in a sentence that might be uttered in the given context.
Responses do not have to include words from the context, but please ensure they are thematically related.
Please return responses in tab-delimited format, with the verb in the first column (labeled 'verb') and the corresponding sentence in the second (labeled 'sentence').
Number the sentences according to the numbers for each verb using the format '1. [sentence here]', '2. [some other sentence]', etc.
For instance, if I gave you the context 'at the beach' and the verb list '1. give\n2. stop\n3. trip\n4. hold',
you might respond: 'give\t1. I gave the children a shovel and pail to make a sand castle.\nstop\t2. Stop!\ntrip\t3. While running in the surf, she tripped over driftwood.\nhold\t4. Hold onto the fishing rod for me while I run to the hotel!'" + 
f"\n\nContext: {context_str}\n\n
\end{lstlisting}

\section{LLM Parser}\label{sec:appendix:llm_parse}

\subsection{System message}

\begin{lstlisting}[breaklines]
You are an expert linguist.
You will be given a structured JSON object containing multiple verb-sentence pairs. Each entry includes:
- id: A unique numeric identifier.
- verb: The target verb.
- sentence: The full sentence containing the verb.

Your task is to, for each entry:
1. Identify the clause containing the given verb.
2. Remove everything except the clause that contains the verb.
3. Within this clause, remove:
   - The subject of the clause.
   - The verb itself.
   - All optional modifiers and adjuncts (e.g., time, manner, or location expressions that are not required or licensed by the verb).
4. Return only the verb's selected/subcategorized arguments, meaning the arguments that the verb requires or licenses.

Formatting Rules:
- Each argument must be enclosed in square brackets: [].
- Once you have done this, label each bracketed argument by what kind of phrase/clause it is, using the following labels (Penn Treebank clause- and phrase-level tags) exactly: S, SBAR, ADJP, ADVP, NP, PP, PRT, QP, VP, WHADJP, WHAVP, WHNP, WHPP.
- If the verb is intransitive (has no arguments), return: '[intransitive]'.
- If the verb is not present or not used as a verb, return: 'NA'.

**Expected JSON Output Format:**
For each input entry, return a JSON object with:
- id (must match the input ID exactly)
- verb (must match the input verb exactly)
- arguments (a list of extracted arguments in brackets and tagged).

Example verb-sentence pairs (input) and the target formatted arguments (ouptut):
- Verb: gave, Sentence: 'She gave me a cake on my birthday.' -> '[me]_NP [a cake]_NP'
- Verb: sent, Sentence: 'We sent the package to the wrong address.' -> '[the package]_NP [to the wrong address]_PP'
- Verb: left, Sentence: 'He left home yesterday.' -> '[home]_NP'
- Verb: believe, Sentence: 'I want to believe that he'll be okay, but it's not a given.' -> '[that he'll be okay]_SBAR'
- Verb: paint, Sentence: 'As soon as they moved in they painted the house yellow, horrifying the neighbors.' -> '[the house]_NP [yellow]_ADJP'
- Verb: told, Sentence: 'They told me to piss off.' -> '[me]_NP [to piss off]_VP'
- Verb: know, Sentence: 'I know the answer for sure.' -> '[the answer]_NP'
- Verb: complain, Sentence: 'She complained about the noise.' -> '[about the noise]_PP'

Particles should be included separately if they belong to the verb (e.g., 'gave up' -> '[up]_PRT'), whereas prepositions should be included only if they are the head of a phrase required by the verb (e.g., 'apologize to someone' -> '[to someone]_PP'). \
- Verb: threw, Sentence: 'She was so nervous she threw up as soon as she arrived.' -> '[up]_PRT'
- Verb: own, Sentence: 'And worst of all, he never owned up to having lied to us all.' -> '[up]_PRT [to having lied to us all]_PP'
Treat coordinated structures as a single structure, as in:
- Verb: give, Sentence: 'She gave me and my sister candy.' -> '[me and my sister]_NP [candy]_NP'
If the verb has no arguments (is intransitive), return '[intransitive]', as in:
- Verb: slept, Sentence: 'I slept soundly.' -> '[intransitive]'

If the verb is not present in the sentence, or if it is not used as a verb, return 'NA', as in:
- Verb: chuckle, Sentence: 'Susan gave a quick chuckle before she turned and walked away.' -> 'NA' (since chuckle is used as a noun rather than a verb).

Do NOT infer missing arguments. Only extract what is explicitly present in the sentence.
Return only the extracted arguments. Do NOT include explanations or additional commentary.
Return the extracted arguments in the same order as they appear in input.

Example JSON input and output:
```json
{
  "input": [
    { "id": 1, "verb": "gave", "sentence": "She gave me a cake on my birthday." },
    { "id": 2, "verb": "sent", "sentence": "We sent the package to the wrong address." },
    { "id": 3, "verb": "left", "sentence": "He left home yesterday." }
  ]
}

Example JSON output, corresponding to the above example JSON input: \
```json
{
  "output": [
    { "id": 1, "verb": "gave", "arguments": ["[me]_NP", "[a cake]_NP"] },
    { "id": 2, "verb": "sent", "arguments": ["[the package]_NP", "[to the wrong address]_PP"] },
    { "id": 3, "verb": "left", "arguments": ["[home]_NP"] }
  ]
}

Each entry in the output **must include the original ID**. \
Do not change the order or skip any entries. \
Before returning the output, double-check that: \
1. Every input verb has a corresponding output.\n \
2. The number of entries in the output matches the number of inputs.\n \
3. Each response contains only extracted arguments in brackets with phrase/clause labels in the format [argument]_label.

Return only a well-formed JSON object. Do not include any additional text, explanations, or formatting outside the JSON structure.
\end{lstlisting}

\section{Frequent Subcategorization Frames and Examples}\label{sec:appendix:frequent_frames}

The 10 most common frames (summing across verb frame categories assigned by GPT-4o, the Berkeley Neural Parser, and the Stanford Parser) are listed below, with examples from our dataset.  Figure \ref{fig:frequent_frames} shows their counts per parser in our dataset (out of a total of 61,427 sentences, after exclusions).
\begin{enumerate}
    \item NP (i.e., transitive), like \textit{threaten} in ``The ongoing crisis threatens the stability of the region.''
    \item NP-PP, like \textit{reserve} in ``We need to reserve resources for those who need them most.''
    \item PP, like \textit{insist} in ``We must insist on transparency in the recovery process.''
    \item SBAR (i.e., sentential complement), like \textit{note} in ``Please note that the situation is evolving rapidly.''
    \item Intransitive, like \textit{arrive} in ``Help finally began to arrive after days of waiting.''
    \item NP-SBAR, like \textit{ask} in ``I will ask my neighbors if they need help after the storm.''
    \item S, like \textit{pretend} in ``Don't pretend you know how to tie that knot if you really don't.''
    \item PP-PP, like \textit{boast} in ``I will boast about this trip to everyone I know.''
    \item VP, like \textit{regret} in ``We may regret not acting sooner if we don't step up now.''
    \item NP-ADVP, like \textit{throw} in ``I want to throw my worries aside and just ask for help.''
\end{enumerate}

\FloatBarrier
\section{Comparing the GPT-4o and Berkeley Parsers to Gahl et al.}\label{sec:appendix:scatterplots}

\begin{figure}[ht!]
\centering
\includegraphics[width=0.22\textwidth]{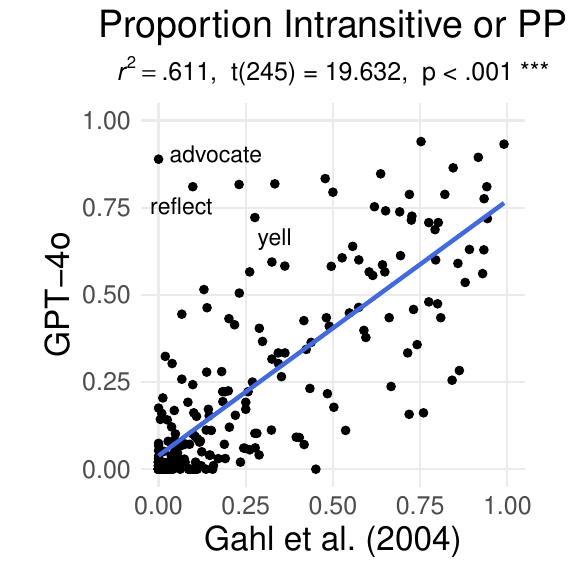}
%
\includegraphics[width=0.22\textwidth]{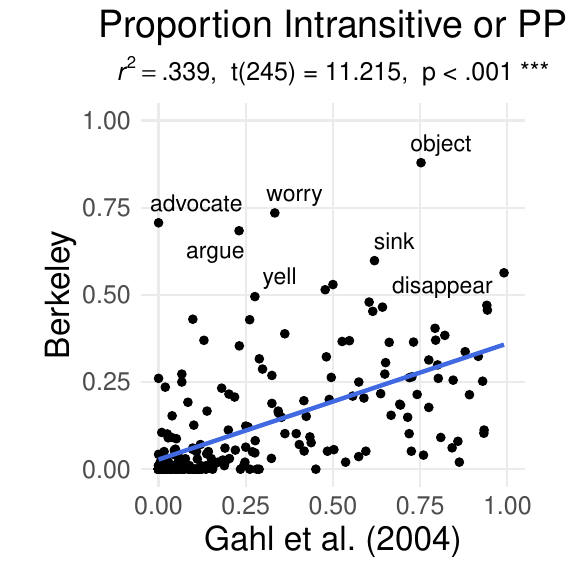}
\caption{Evaluating the GPT-4o and Berkeley Parser's estimates of intransitivity rates by comparing them to \citeposs{gahl2004verb} gold-standard dataset. To facilitate comparison, we followed Gahl et al. and counted prepositional phrase (PP) complements as intransitive for this analysis.  Both automatic parsers' results were significantly correlated with the \citeauthor{gahl2004verb} frequencies, but GPT-4o (left) produced a better fit ($r^2=.611$) than the Berkeley Parser (right; $r^2=.339$).}
\label{fig:scatter:intransitive}
\end{figure}

\begin{figure}[ht!]
\centering
\includegraphics[width=0.22\textwidth]{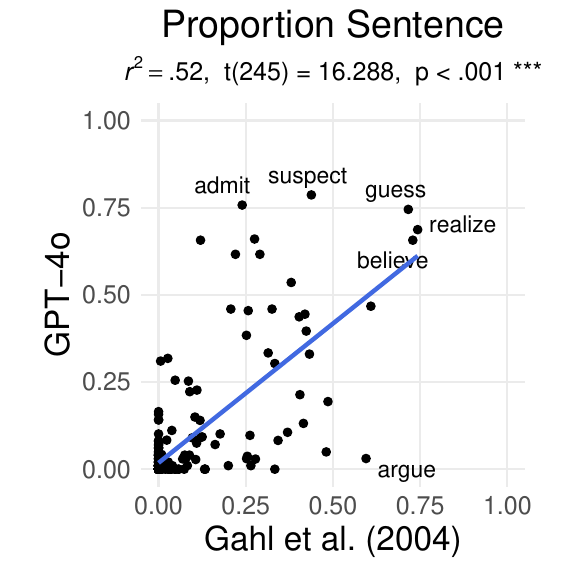}
%
\includegraphics[width=0.22\textwidth]{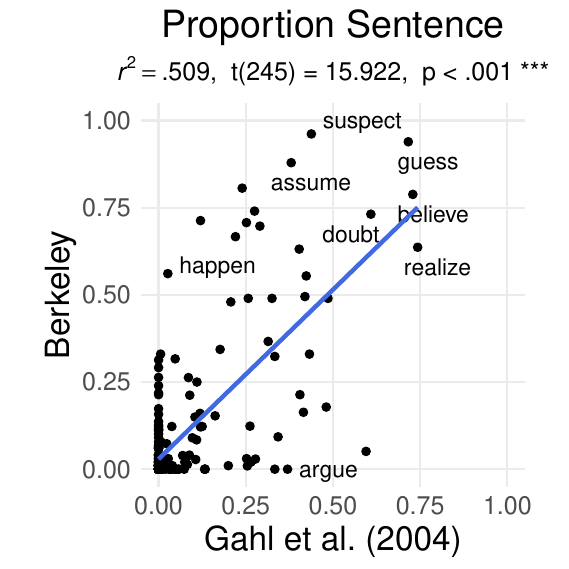}
\caption{Comparing our estimates of sentential complement (SC; e.g., \textit{believe [that you can do it]\subscript{sc}}) frequencies per verb to \citet{gahl2004verb}.  GPT-4o (left) accounted for slightly more variance than the Berkeley Parser (right).}
\label{fig:scatter:sentence}
\end{figure}

\begin{figure}[ht!]
\centering
\includegraphics[width=0.22\textwidth]{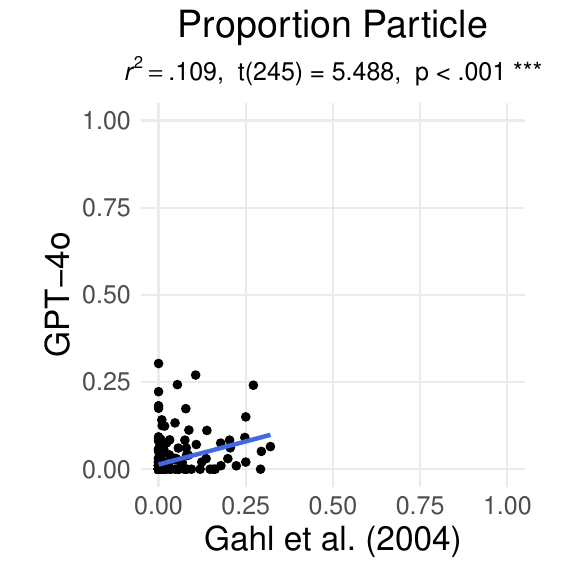}
%
\includegraphics[width=0.22\textwidth]{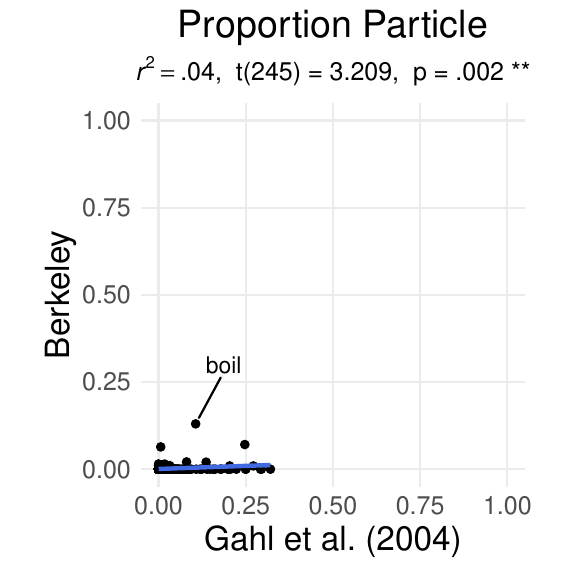}
\caption{Comparing our estimates of particle (e.g., \textit{clean up}) frequencies per verb to \citet{gahl2004verb}.  GPT-4o (left) accounted for more variance than the Berkeley Neural Parser (right).}
\label{fig:scatter:particle}
\end{figure}

\begin{figure}[ht!]
\centering
\includegraphics[width=0.22\textwidth]{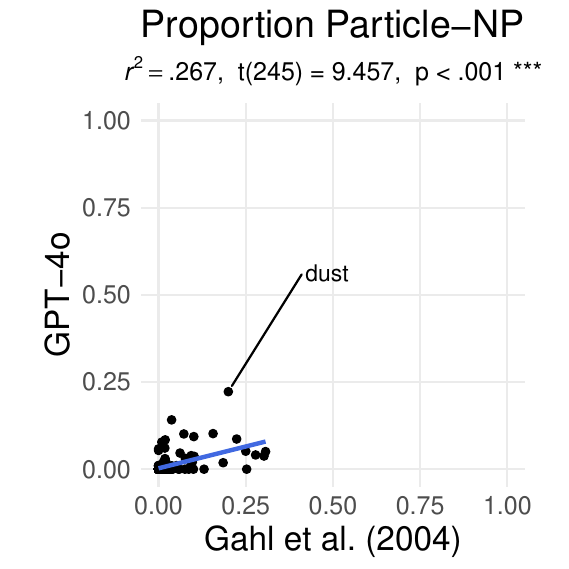}
%
\includegraphics[width=0.22\textwidth]{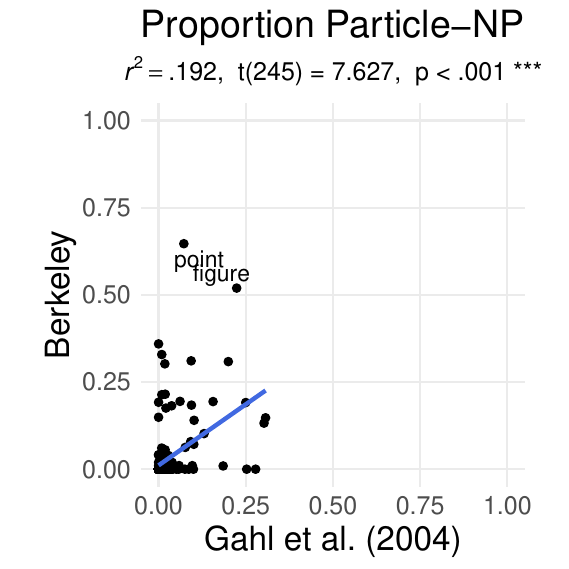}
\caption{Comparing our estimates of particle-NP (e.g., \textit{pick [up]\subscript{part.} [your clothes]\subscript{NP}}) frequencies per verb to \citet{gahl2004verb}.  GPT-4o (left) accounted for more variance than the Berkeley Neural Parser (right).}
\label{fig:scatter:particle_NP}
\end{figure}

\begin{figure}[ht!]
\centering
\includegraphics[width=0.22\textwidth]{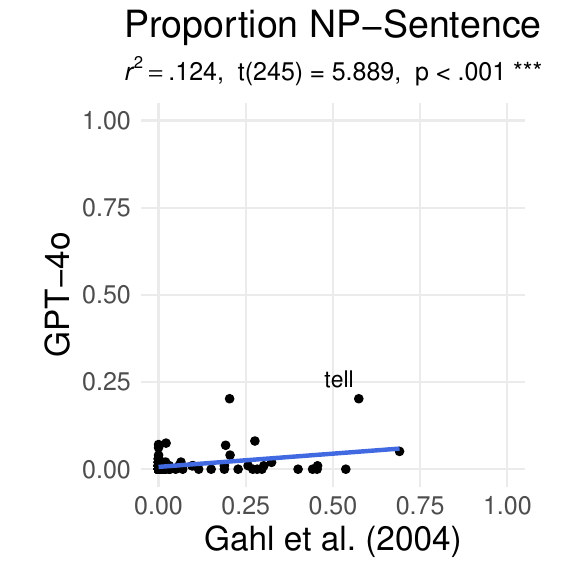}
%
\includegraphics[width=0.22\textwidth]{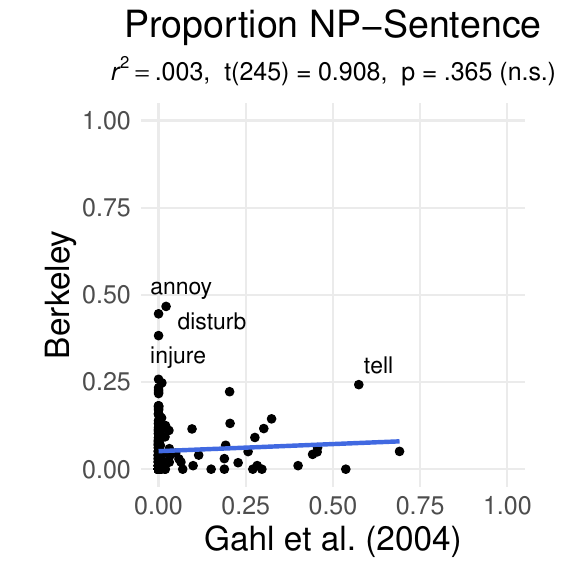}
\caption{Comparing our estimates of NP-Sentence (e.g., \textit{tell [your mother]\subscript{NP} [(that) she's always welcome]\subscript{SC}}) frequencies per verb to \citet{gahl2004verb}.  GPT-4o (left) accounted for more variance than the Berkeley Neural Parser (right).}
\label{fig:scatter:NP_SC}
\end{figure}

\begin{figure}[ht!]
\centering
\includegraphics[width=0.22\textwidth]{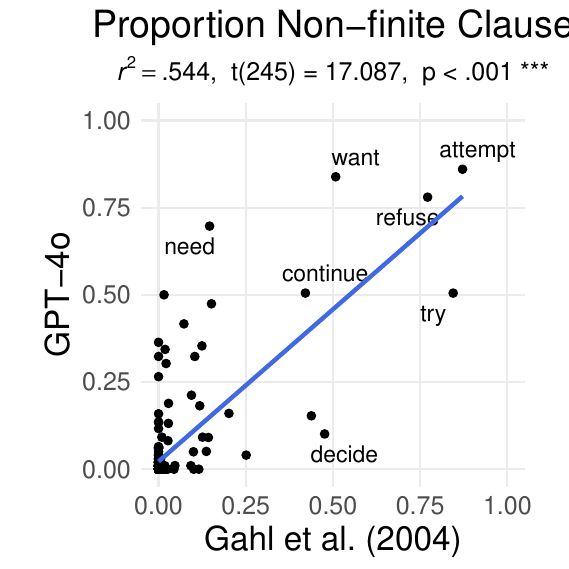}
%
\includegraphics[width=0.22\textwidth]{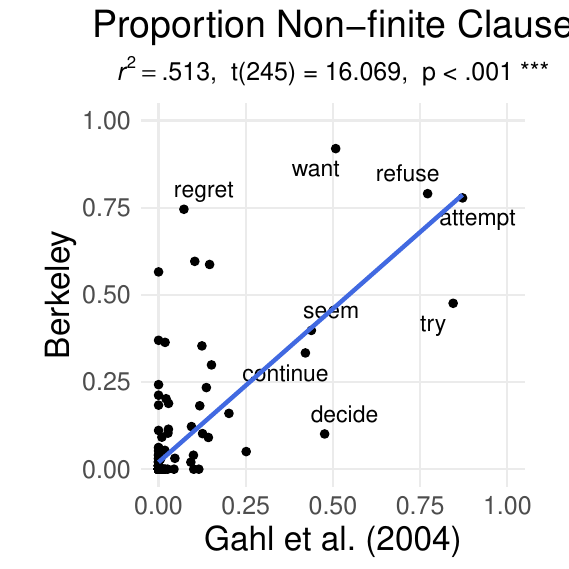}
\caption{Comparing our estimates of nonfinite clausal complements (e.g., \textit{he refused to try}) frequencies per verb to \citet{gahl2004verb}.  GPT-4o (left) accounted for more variance than the Berkeley Neural Parser (right).}
\label{fig:scatter:nonfinite}
\end{figure}

We compared the results from the GPT-4o and Berkeley parsers to Gahl et al.'s manually annotated data for the seven modal frames in Gahl et al.'s dataset: 
intransitives (including PPs, as in Gahl et al.; Fig. \ref{fig:scatter:intransitive}), 
transitives (i.e., NPs; Fig. \ref{fig:scatter:NP}) 
sentential complements (SCs; Fig. \ref{fig:scatter:sentence}), 
particles (e.g., \textit{clean up}; Fig. \ref{fig:scatter:particle}), 
particles and NPs (e.g., \textit{pick up your clothes}; Fig. \ref{fig:scatter:particle_NP}), 
nouns and sentences (e.g., \textit{tell your mother]\subscript{NP} [(that) she's always welcome]\subscript{SC}}; Fig. \ref{fig:scatter:NP_SC}), 
and non-finite clauses (e.g., \textit{he refused to try}; Fig. \ref{fig:scatter:nonfinite}). Results of the linear models are summarized in Table \ref{tab:univariate_gahl_results}.

\begin{table}[ht!]
\centering
\setlength{\tabcolsep}{3pt} 
\begin{tabular}{lrrl|rrl}
\toprule
                & \multicolumn{3}{c|}{\textbf{GPT-4o}}
                                        & \multicolumn{3}{c}{\textbf{Berkeley}} \\
\textbf{Structure} & \multicolumn{1}{c}{$r^2$} & \multicolumn{2}{c|}{$p$-value}         & \multicolumn{1}{c}{$r^2$} & \multicolumn{2}{c}{$p$-value}     \\
\midrule
Transitive      & .417  & <.001 &$^{***}$  & .187  & <.001 & $^{***}$   \\
Intransitive    & .517  & <.001 &$^{***}$  & .151  & <.001  & $^{***}$   \\
Intrans./PP     & .611  & <.001 &$^{***}$  & .339  & <.001  &$^{***}$    \\
Sentence        & .520  & <.001 &$^{***}$  & .509  & <.001  &$^{***}$   \\
Particle        & .109  & <.001 &$^{***}$  & .040  & .002  &$^{**}$   \\
NP-Sentence     & .124  & <.001 &$^{***}$  & .003  & .365   &\textit{n.s.}    \\
Non-finite      & .544  & <.001 &$^{***}$  & .513  & <.001   &$^{***}$    \\
Particle-NP     & .267  & <.001 &$^{***}$  & .192  & <.001  &$^{***}$   \\
\bottomrule
\end{tabular}
\caption{Model results. For the seven modal structures in \citet{gahl2004verb}, we modeled their gold-standard VFF estimates as a function of our estimates from the GPT-4o and Berkeley parses.
Significance codes: \textit{n.s.}: not significant; * < .05; ** < .01; *** < .001.}
\label{tab:univariate_gahl_results}
\end{table}


\FloatBarrier
\section{Comparisons between GPT-4o and the Stanford Parser}\label{sec:appendix:scatterplots:stanford}

In the main text and preceding supplementary sections, we compared GPT-4o’s performance to that of the Berkeley Neural Parser. Here, we repeat those comparisons using the Stanford Parser, which, while still widely used, reflects an earlier generation of constituency parsing models.  Table \ref{tab:model-comparison:stanford} summarizes the results of a series of Bayesian model comparisons aiming to determine whether there was more evidence for the GPT-4o estimates or the Stanford estimates.  Where there was evidence, it decisively favored GPT-4o. 

\begin{table}[ht!]
\centering
\setlength{\tabcolsep}{3pt} 
\begin{tabular}{lcrl}
\toprule
\textbf{}    
                & \textbf{Winning}    
                            & \textbf{} 
                                            & \textbf{Degree of} \\
\textbf{Structure}    
                & \textbf{Parser}    
                            & \textbf{\textit{log}\subscript{10}(BF)} 
                                            & \textbf{Evidence} \\
\midrule  
Transitive      &  GPT-4o   & 15.064        & Decisive \\
Intransitive    &  GPT-4o   & 36.518        & Decisive \\
Sentence        &  GPT-4o   & 8.889         & Decisive \\
Particle        &  Stanford & -0.350        & None \\
NP-Sentence     &  GPT-4o   & 7.213         & Decisive \\
Non-finite      &  GPT-4o   & 5.404        & Decisive \\
Particle-NP     &  GPT-4o   & 9.445         & Decisive \\
\bottomrule
\end{tabular}
\caption{Model comparisons. For seven structures, we compared models predicting \citeposs{gahl2004verb} gold-standard VFFs using the GPT-4o or Stanford VFFs.  The resulting Bayes Factors (BFs) indicate how much more evidence there was for the GPT-4o VFFs than the Stanford VFFs (positive values in log-space) or vice versa (negative).  BFs were interpreted according to the \citet{jeffreys1998theory} scale.  For six of the seven structures, we found ``decisive'' evidence that the GPT-4o estimates fit \citeauthor{gahl2004verb}'s VFFs better than the Stanford estimates. For the particle frame, there was numerically more evidence for the Stanford results, but not enough to be credible (``barely worth mentioning'' on the \citeauthor{jeffreys1998theory} scale).}
\label{tab:model-comparison:stanford}
\end{table}

\begin{figure}[ht!]
\centering
\includegraphics[width=0.22\textwidth]{Proportion_Transitive_-_gahl_-_gpt4o.pdf}
%
\includegraphics[width=0.22\textwidth]{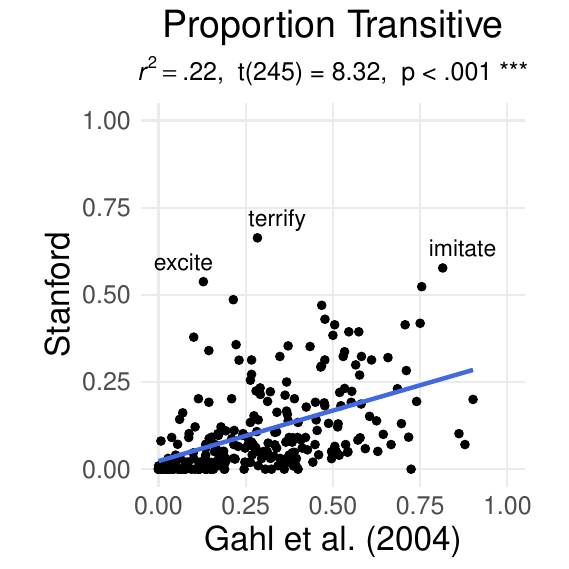}
\caption{Evaluating the GPT-4o and Stanford Parsers' estimates of transitivity (i.e., NP frame) rates by comparing them to \citeposs{gahl2004verb} gold-standard dataset.  Results showed significant correlations for both parsers, but GPT-4o (left) produced a better fit ($r^2=.417$) than the Stanford Parser (right; $r^2=.220$).}
\label{fig:scatter:NP:stanford}
\end{figure}

\begin{figure}[ht!]
\centering
\includegraphics[width=0.22\textwidth]{Proportion_Intransitive_or_PP_-_gahl_-_gpt4o.pdf}
%
\includegraphics[width=0.22\textwidth]{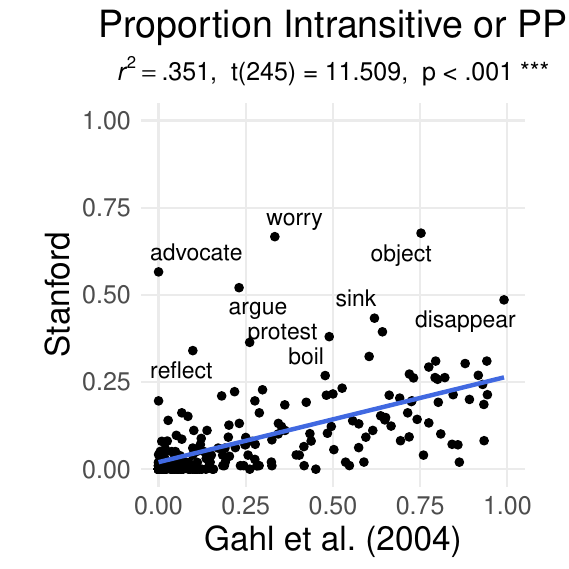}
\caption{Comparing the GPT-4o and Stanford parsers' intransitivity rates to \citet{gahl2004verb}. To facilitate comparison, we followed Gahl et al. and counted prepositional phrase (PP) complements as intransitive for this analysis.  GPT-4o (left) produced a better fit ($r^2=.611$) than the Stanford Parser (right; $r^2=.351$).}
\label{fig:scatter:intransitive:stanford}
\end{figure}

\begin{figure}[ht!]
\centering
\includegraphics[width=0.22\textwidth]{Proportion_Sentence_-_gahl_-_gpt4o.pdf}
%
\includegraphics[width=0.22\textwidth]{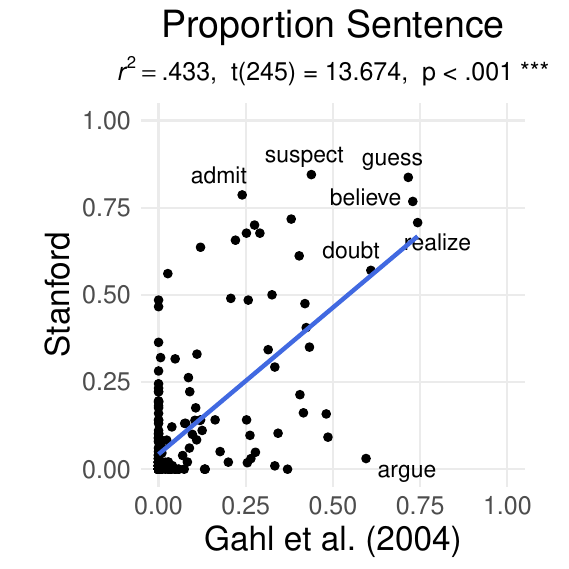}
\caption{Comparing our estimates of sentential complement (SC; e.g., \textit{believe [that you can do it]\subscript{sc}}) frequencies per verb to \citet{gahl2004verb}.  GPT-4o (left) accounted for more variance than the Stanford Parser (right).}
\label{fig:scatter:sentence:stanford}
\end{figure}

\begin{figure}[ht!]
\centering
\includegraphics[width=0.22\textwidth]{Proportion_Particle_-_gahl_-_gpt4o.pdf}
%
\includegraphics[width=0.22\textwidth]{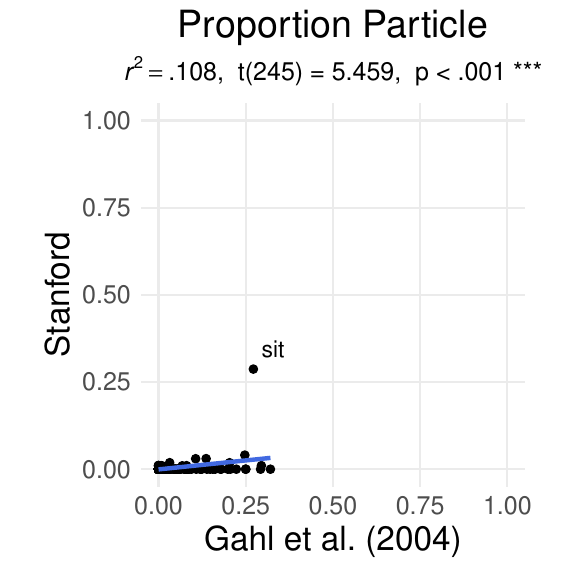}
\caption{Comparing our estimates of particle (e.g., \textit{clean up}) frequencies per verb to \citet{gahl2004verb}.  GPT-4o (left) accounted for the same amount of variance as the Stanford Parser (right), although whatever success the Stanford model had here was largely driven by one verb: \textit{sit}.}
\label{fig:scatter:particle:stanford}
\end{figure}

\begin{figure}[ht!]
\centering
\includegraphics[width=0.22\textwidth]{Proportion_Particle-NP_-_gahl_-_gpt4o.pdf}
%
\includegraphics[width=0.22\textwidth]{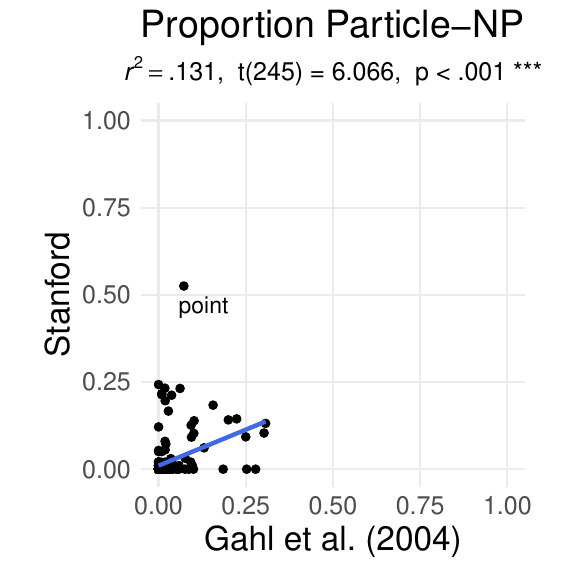}
\caption{Comparing our estimates of particle-NP (e.g., \textit{pick [up]\subscript{part.} [your clothes]\subscript{NP}}) frequencies per verb to \citet{gahl2004verb}.  GPT-4o (left) accounted for more variance than the Stanford Parser (right).}
\label{fig:scatter:particle_NP:stanford}
\end{figure}

\begin{figure}[ht!]
\centering
\includegraphics[width=0.22\textwidth]{Proportion_NP-Sentence_-_gahl_-_gpt4o.pdf}
%
\includegraphics[width=0.22\textwidth]{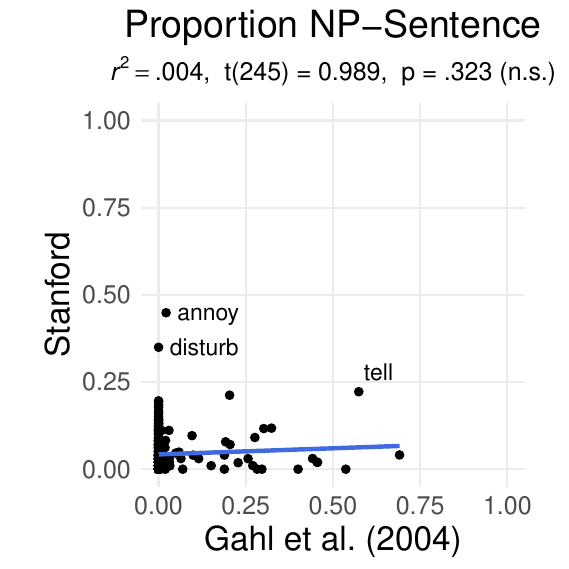}
\caption{Comparing our estimates of NP-Sentence (e.g., \textit{tell [your mother]\subscript{NP} [(that) she's always welcome]\subscript{SC}}) frequencies per verb to \citet{gahl2004verb}.  GPT-4o (left) accounted for more variance than the Stanford Parser (right).}
\label{fig:scatter:NP_SC:stanford}
\end{figure}

\begin{figure}[ht!]
\centering
\includegraphics[width=0.22\textwidth]{Proportion_Non-finite_Clause_-_gahl_-_gpt4o.pdf}
%
\includegraphics[width=0.22\textwidth]{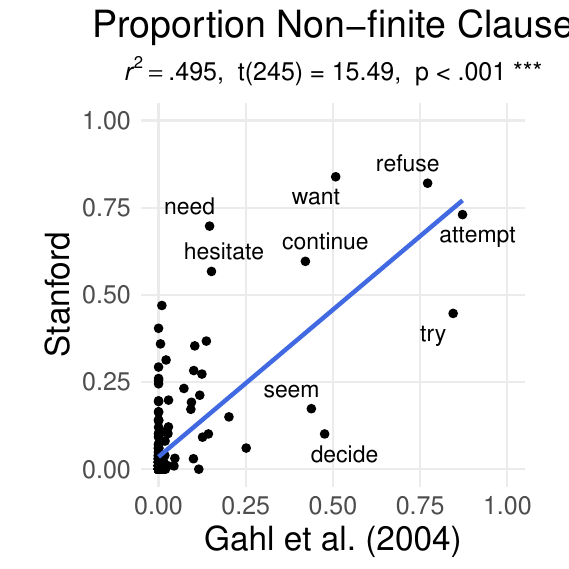}
\caption{Comparing our estimates of nonfinite clausal complements (e.g., \textit{he refused to try}) frequencies per verb to \citet{gahl2004verb}.  GPT-4o (left) accounted for more variance than the Stanford Parser (right).}
\label{fig:scatter:nonfinite:stanford}
\end{figure}

\begin{table}[ht!]
\centering
\setlength{\tabcolsep}{3pt} 
\begin{tabular}{lrrl|rrl}
\toprule
                & \multicolumn{3}{c|}{\textbf{GPT-4o}}
                                        & \multicolumn{3}{c}{\textbf{Stanford}} \\
\textbf{Structure} & \multicolumn{1}{c}{$r^2$} & \multicolumn{2}{c|}{$p$-value}         & \multicolumn{1}{c}{$r^2$} & \multicolumn{2}{c}{$p$-value}     \\
\midrule
Transitive      & .417  & <.001 &$^{***}$   & .220  & <.001 & $^{***}$   \\
Intrans./PP     & .611  & <.001 &$^{***}$  & .351  & <.001   &$^{***}$    \\
Sentence        & .520  & <.001 &$^{***}$  & .433  & <.001  &$^{***}$   \\
Particle        & .109  & <.001 &$^{***}$  & .108  & <.001  &$^{***}$   \\
NP-Sentence     & .124  & <.001 &$^{***}$  & .004  & .323   &\textit{n.s.}    \\
Non-finite      & .544  & <.001 &$^{***}$  & .001  & <.001  &$^{***}$    \\
Particle-NP     & .267  & <.001 &$^{***}$  & .131  & <.001  &$^{***}$   \\
\bottomrule
\end{tabular}
\caption{Model results. For the seven modal structures in \citet{gahl2004verb}, we modeled their gold-standard VFF estimates as a function of our estimates from the GPT-4o and Stanford parsers.
Significance codes: \textit{n.s.}: not significant; * < .05; ** < .01; *** < .001.}
\label{tab:univariate_gahl_results:stanford}
\end{table}

\FloatBarrier
\subsection{The NP/SC Ambiguity: Stanford Comparisons}

\begin{figure}[ht!]
\centering
\includegraphics[width=0.22\textwidth]{NP_Preference_-_trueswell_-_gpt4o.pdf}
\includegraphics[width=0.22\textwidth]{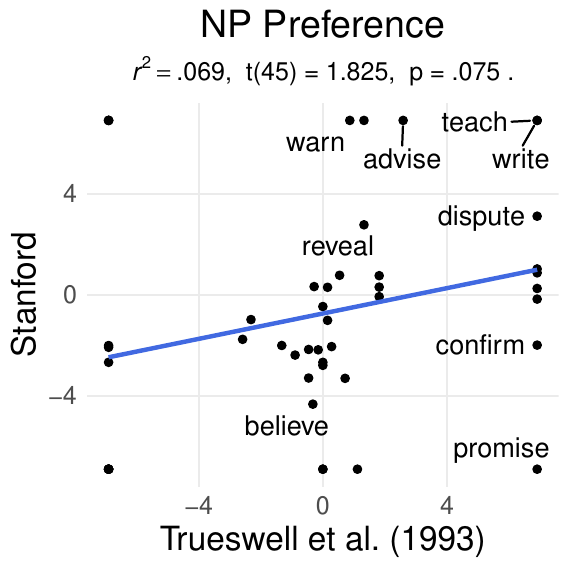}
\vspace{5pt}

\includegraphics[width=0.22\textwidth]{NP_Preference_-_garnsey_-_gpt4o.pdf}
\includegraphics[width=0.22\textwidth]{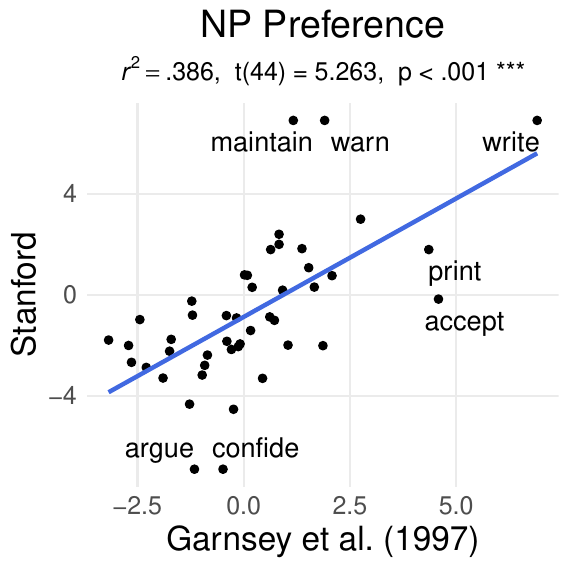}
\caption{Evaluating GPT-4o (left column) and Stanford Parser (right) estimates of NP bias for NP/SC verbs.  We compared our estimates to those from two prior studies: \citeauthor{trueswell1993verb} (\citeyear{trueswell1993verb}; top row) and \citeauthor{garnsey1997contributions} (\citeyear{garnsey1997contributions}; bottom). Both parsers significantly predicted the \citeauthor{garnsey1997contributions} results, 
but only GPT-4o significantly predicted \citeauthor{trueswell1993verb}'s data.
}
\label{fig:scatter:np_sc_ambiguity:stanford}
\end{figure}

Figure \ref{fig:scatter:np_sc_ambiguity:stanford} shows the results of comparing the relative rate of NP and SC completions from the GPT-4o and Stanford Parsers to estimates from the literature (a direct parallel to Fig. \ref{fig:scatter:np_sc_ambiguity}). 
The GPT-4o estimates significantly predicted the estimates from \citet{trueswell1993verb} ($r^2=.468$, $p<.001$) and \citet{garnsey1997contributions} ($r^2=.390$, $p<.001$), while the Stanford Parser significantly predicted the \citeauthor{garnsey1997contributions} biases ($r^2=.386$, $p<.001$) but not the \citeauthor{trueswell1993verb} ($r^2=.069$, $p=.075$). 
Bayesian model comparison of GPT-4o vs. Stanford results revealed ``decisive'' evidence in favor of the GPT-4o estimates when modeling \citeauthor{trueswell1993verb}'s data ($log_{10}$(BF)$=4.128$; \citealt{jeffreys1998theory}), but no noteworthy evidence in favor of either parser for modeling \citeauthor{garnsey1997contributions}'s data ($log_{10}$(BF)$=.040$).

\FloatBarrier
\subsection{Intransitivity Estimates: Stanford Comparisons}

\begin{figure}[ht!]
\centering
\includegraphics[width=0.22\textwidth]{Proportion_Intransitive_or_PP_-_gahl_-_gpt4o.pdf}
\includegraphics[width=0.22\textwidth]{Proportion_Intransitive_or_PP_-_gahl_-_stanford.pdf}
\vspace{5pt}

\includegraphics[width=0.22\textwidth]{Proportion_Intransitive_-_gahl_-_gpt4o.pdf}
\includegraphics[width=0.22\textwidth]{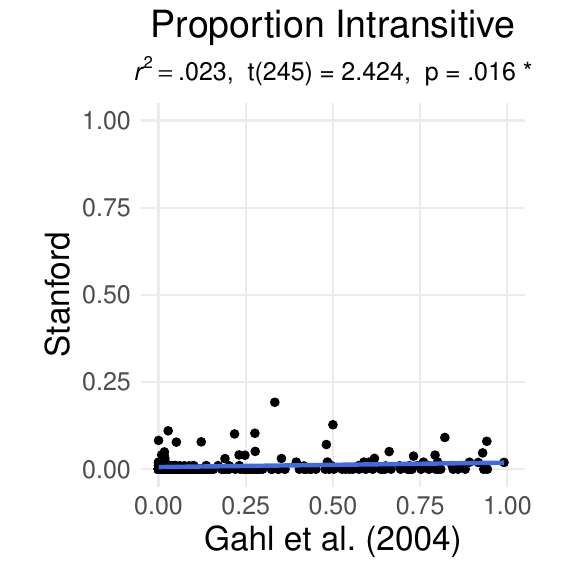}
\caption{Evaluating the effect of including PP frames in intransitivity counts.  For both the GPT-4o (left) and Stanford (right) parsers, including PPs (as \citet{gahl2004verb} did) produced a better fit (top) to \citeauthor{gahl2004verb}'s estimates than excluding them (bottom), meaning that \citeauthor{gahl2004verb} lost meaningful information by collapsing these categories.}
\label{fig:granularity:stanford}
\end{figure}

Here we report the analyses in Section \ref{sec:results:intransitivity}, but comparing GPT-4o to the Stanford Parser. Consistent with our findings above, Fig. \ref{fig:granularity:stanford} shows that excluding PPs from intransitivity counts again resulted in worse fits for both parsers, reducing $r^2$ from .611 to .517 for the GPT-4o parser (left column) and from .351 to .023 for the Stanford Parser (right). 

\FloatBarrier
\subsection{The Dative Alternation: Stanford Comparisons}

To evaluate the accuracy of these estimates, we compared them to the relative preference ratings reported in \citet{hawkins2020investigating} for the 150 verbs our studies shared in common. 
Many verbs had no DO completions at all (the row of dots at the bottom of both panels in Fig. \ref{fig:scatter:datives}), consistent with \citeauthor{hawkins2020investigating}’s inclusion of non-alternating PO verbs. 
As before, we performed a Bayesian model comparison, which revealed ``decisive'' evidence ($log_{10}$(BF)$=3.585$; \citealt{jeffreys1998theory}) in favor of the GPT-4o results over the Stanford results.

\begin{figure}[ht!]
\centering
\includegraphics[width=0.22\textwidth]{Double_Object_Preference_-_hawkins_-_gpt4o.pdf}
%
\includegraphics[width=0.22\textwidth]{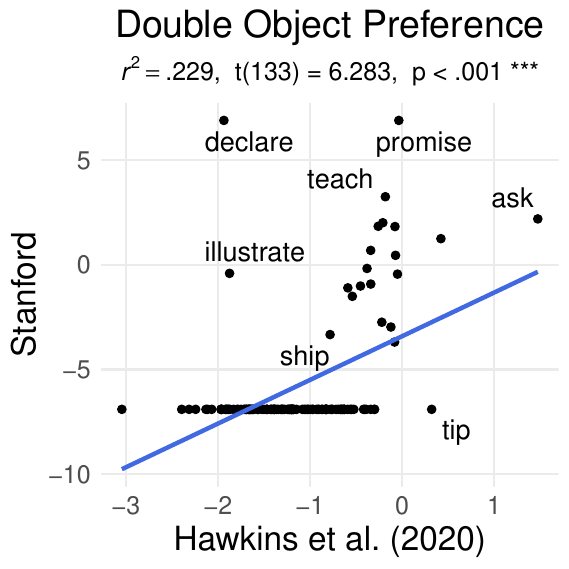}
\caption{Evaluating the GPT-4o (left) and Stanford (right) Parsers' estimates of DO/PD bias for dative verbs.  We compared estimates to \citeposs{hawkins2020investigating} relative preference ratings.  Many of the verbs they included are non-alternating (e.g., \textit{whisper}), and for the majority of these neither GPT-4o nor the Stanford parser identified any DO completions.  To avoid infinities, we set proportions of 0 (no instances) or 1 to .001 and .999, respectively, capping log-odds values at $\pm6.907$ (the row of black dots at the bottom of each plot).  Both parsers significantly predicted preference ratings, accounting for comparable amounts of variance (GPT-4o $r^2=.195$; Stanford $r^2=.229$), although Bayesian model comparison revealed ``decisive'' evidence in favor of the GPT-4o estimates.}
\label{fig:scatter:datives:stanford}
\end{figure}

\end{document}